\documentclass[10pt,twocolumn,letterpaper]{article}
\usepackage{cvpr}      
\usepackage{graphicx}
\usepackage{amsmath}
\usepackage{amssymb}
\usepackage{booktabs}
\usepackage{bbding} 
\usepackage{cite}
\usepackage{appendix}
\usepackage[pagebackref,breaklinks,colorlinks]{hyperref}
\usepackage{multirow}
\usepackage[table,xcdraw]{xcolor}
\usepackage{booktabs}
\usepackage{flushend}
\usepackage[accsupp]{axessibility}
\usepackage[capitalize]{cleveref}
\crefname{section}{Sec.}{Secs.}
\Crefname{section}{Section}{Sections}
\Crefname{table}{Table}{Tables}
\crefname{table}{Tab.}{Tabs.}

\def\cvprPaperID{6126} 
\def\confName{CVPR}
\def\confYear{2022}

\begin{document}

\title{Dynamic Prototype Convolution Network for Few-Shot Semantic Segmentation}
\author{Jie Liu$^{1}$\thanks{Equal contribution.}~, Yanqi Bao$^{2*}$, Guo-Sen Xie$^{3,4}$\thanks{Corresponding author.}~, Huan Xiong$^4$, Jan-Jakob Sonke$^{5}$, Efstratios Gavves$^1$\\
{\normalsize{$^1$University of Amsterdam, Netherlands}}~~~{\normalsize{$^2$Northeastern University, China}}~~~{\normalsize{$^5$The Netherlands Cancer Institute, Netherlands}}\\{\normalsize{$^3$Nanjing University of Science and Technology, China}}~~~{\normalsize{$^4$Mohamed bin Zayed University of Artificial Intelligence, UAE}}}
\maketitle
\begin{abstract}
The key challenge for few-shot semantic segmentation (FSS) is how to tailor a desirable interaction among support and query features and/or their prototypes, under the episodic training scenario. Most existing FSS methods implement such support/query interactions by solely leveraging \it{plain} operations -- e.g., cosine similarity and feature concatenation -- for segmenting the query objects. However, these interaction approaches usually cannot well capture the intrinsic object details in the query images that are widely encountered in FSS, e.g., if the query object to be segmented has holes and slots, inaccurate segmentation almost always happens. To this end, we propose a dynamic prototype convolution network (DPCN) to fully capture the aforementioned intrinsic details for accurate FSS. Specifically, in DPCN, a dynamic convolution module (DCM) is firstly proposed to generate dynamic kernels from support foreground, then information interaction is achieved by convolution operations over query features using these kernels. Moreover, we equip DPCN with a support activation module (SAM) and a feature filtering module (FFM) to generate pseudo mask and filter out background information for the query images, respectively. SAM and FFM together can mine enriched context information from the query features. 
Our DPCN is also flexible and efficient under the k-shot FSS setting. Extensive experiments on PASCAL-$5^i$ and COCO-$20^i$ show that DPCN yields superior performances under both 1-shot and 5-shot settings.
\end{abstract}
\section{Introduction}
\label{Introduction}
Semantic segmentation has achieved tremendous success due to the advancement of deep convolutional neural networks \cite{he2016deep, huang2017densely,simonyan2014very}. Nevertheless, most leading image semantic segmentation models rely on large amounts of training images with pixel-wise annotations, which require huge human efforts. 
Semi- and weakly-supervised segmentation methods \cite{ luo2020semi, sun2020mining, wang2020self} are proposed to alleviate such expensive annotation cost. However, both semi- and weakly-supervised methods have to face a significant performance drop when only a few annotated samples for a novel object are available. In such a case, few-shot semantic segmentation (FSS) \cite{shaban2017one} is introduced to allow for dense pixel-wise prediction on novel object categories given only a few annotated samples.

\begin{figure}[!t]
	\centering
	\includegraphics[width=\columnwidth]{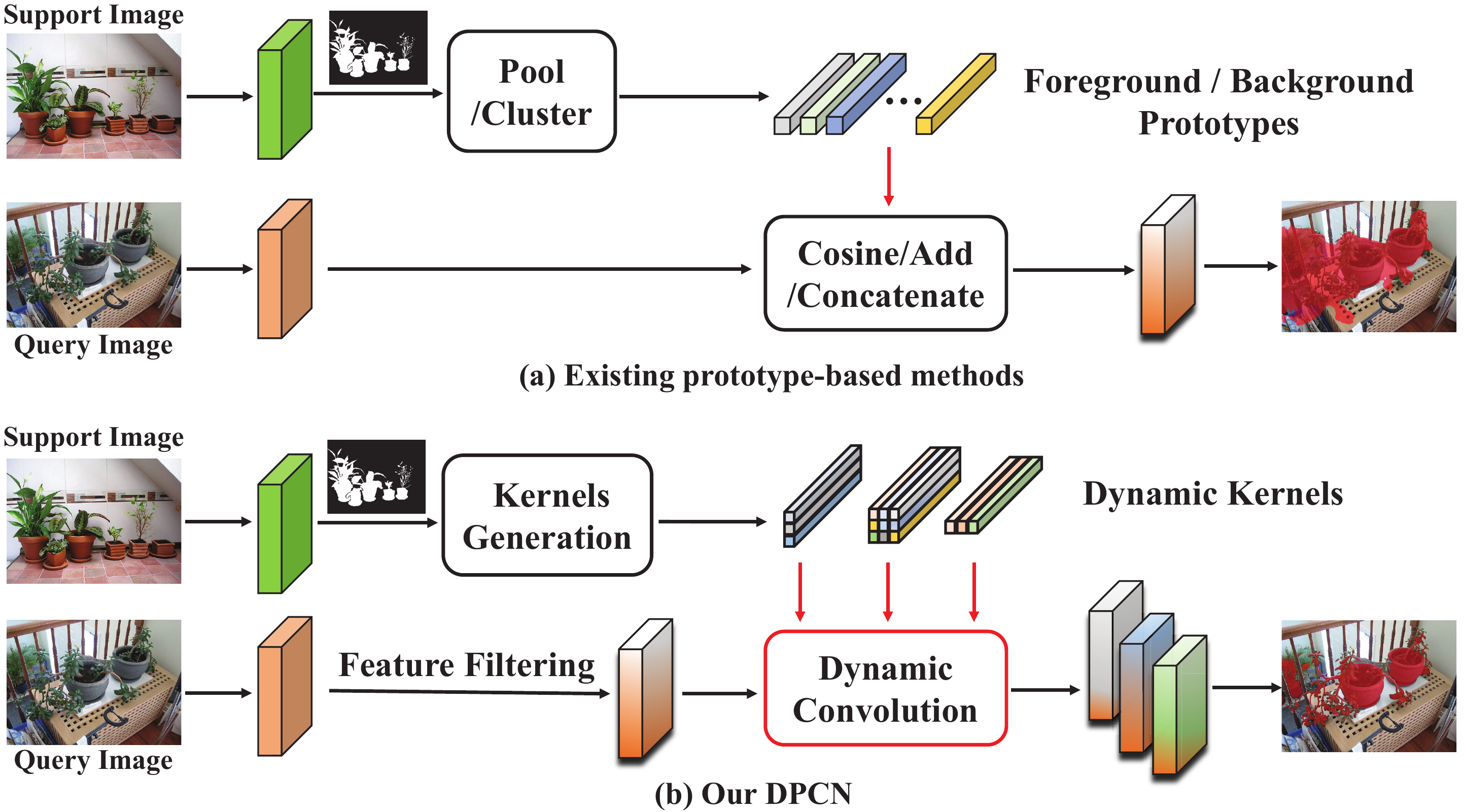}
	\caption{\textbf{Comparison between (a) existing prototype-based methods and (b) proposed DPCN. (a) Existing methods} usually adopt mask average pooling or clustering on support foreground to get multiple foreground/background prototypes. Then the information interaction between support and query can be plain operations, e.g., cosine similarity, element-wise sum, and channel-wise concatenation. However, this paradigm fails to segment the intrinsic details of \emph{plants}, because this insufficient interaction cannot well address the appearance and shape variations (e.g., holes and slots in the \emph{plants}) in FSS. \textbf{(b) Our DPCN} can well segment \emph{plants} and capture the intrinsic subtle details. This benefits from dynamic convolution on query features with dynamic kernels generated from the support foreground features.}
	\label{fig1}
\end{figure}

Typically, most FSS methods adopt an episode-based meta-learning strategy \cite{xie2021scale}, and each episode is composed of the support set and the query set, which share the same object class. The support set contains a few support images with pixel-wise annotation.
FSS models are expected to learn to predict the segmentation mask for images in the query set with the guidance of the support set.
Learning is based on episodes available with annotations during training.
At test time, the model is expected to segment a query image with respect to a class of interest, again provided with corresponding support set, only this time the class of interest in the query and support set is novel and not previously seen.

Currently, the most leading FSS methods are the prototype-based ones \cite{tian2020prior,Li_2021_CVPR}. As in Fig.~\ref{fig1}(a), prototype-based paradigm typically generates multiple foreground and/or background prototypes by utilizing mask average pooling and/or clustering over support features. These prototypes are supposed to contain representative information of the target object in the support images, thus their interactions with query features by cosine similarity, element-wise summation, and feature concatenation can produce necessary predictions for the objects in the query image. However, the predictions achieved by solely relying on such limited prototypes and {\it{plain}} operations are inevitably losing some intrinsic object details in the query image. For instance, as in Fig.~\ref{fig1}(a), since the objects {\it plants} have holes and slots which are intrinsic details, the segmented objects cannot well cover these details, i.e., a defective over-segmentation is achieved in this case. In addition, in the presence of large object variations (e.g., appearance and scale) in FSS, it is usually difficult to comprehensively encode the adequate patterns of the target objects by solely considering the support information as in most previous prototype-based methods.

To address the above challenges, we propose a dynamic prototype convolution network (DPCN) to fully capture the intrinsic object details for accurate FSS. DPCN belongs to prototype-based methods yet with several elegant extensions and merits. Specifically, we first propose a dynamic convolution module (DCM) to achieve more adequate interaction between support and query features, thus leading to more accurate prediction for the query objects. As in Fig. \ref{fig1}(b), we leverage three dynamic kernels, i.e., a square kernel and two asymmetric kernels, generated from the support foreground features. Then three convolution operations are employed in parallel onto the query features using these dynamic kernels. This interaction strategy is simple yet important to comprehensively tackle large object variations (e.g., appearance and scales) and can capture the intrinsic object details. Intuitively, the square kernel is capable of capturing the main information of an object (e.g., main body of the {\it plant} in Fig. \ref{fig1}(b)); By contrast, asymmetric kernels (i.e., kernel with size $\textit{d}\times1$ or $1\times\textit{d}$ aim to capture the subtle object details, e.g., leaves in Fig.~\ref{fig1}(b). As such, DPCN equipped with DCM can better handle the intrinsic object details using an extremely simple way.

Moreover, to comprehensively encode the adequate patterns of the target objects, we propose a support activation module (SAM) and a feature filtering module (FFM) to mine as much object-related context information from query image as possible. Specifically, SAM generates support activation maps and initial pseudo query mask using high-level support and query features. Then the support prototypes and pseudo query foreground features are fused to generate a refined pseudo mask for the query image in FFM. Compared with the original pseudo query mask, the refined one contains more object foreground context while filtering some noise information. Therefore, rich object-related context information from both support and query images are aggregated to the final feature, leading to better segmentation performance. Our main contributions are as follows:

$\bullet$ We propose a dynamic prototype convolution network (DPCN) to capture the intrinsic object details for accurate FSS. To the best of knowledge, we are the first one to do this in the FSS domain.

$\bullet$ We propose a novel dynamic convolution module (DCM) to achieve adequate support-query interactions. DCM can serve as a plug-and-play component to improve existing prototype learning methods.

$\bullet$ We propose a support activation module (SAM) and a feature filtering module (FFM) to mine complementary information of target objects from query images.

\section{Related work}
\label{Relatedwork}



\subsection{Semantic Segmentation}
Semantic segmentation is a classical computer vision task which aims to give pixel-wise prediction for an input image. Recently, various networks \cite{long2015fully} have been actively designed to further improve the semantic segmentation results. For capturing more contextual informations, dilated convolution \cite{yu2015multi}, pyramid pooling \cite{zhao2017pyramid}, and deformable convolution \cite{dai2017deformable}, are proposed to enlarge the receptive filed. Meanwhile, some models leverage attention mechanisms~\cite{zhang2018context,yuan2020object,fu2019dual,wang2018non} to capture long-distance dependencies for semantic segmentation, which reach state-of-the-art performances. However, these semantic segmentation approaches still fail to preserve their initial performances when insufficient training data is provided.

\begin{figure*}[!t]
	\vspace{-5mm}
	\centering
	\includegraphics[width=\linewidth]{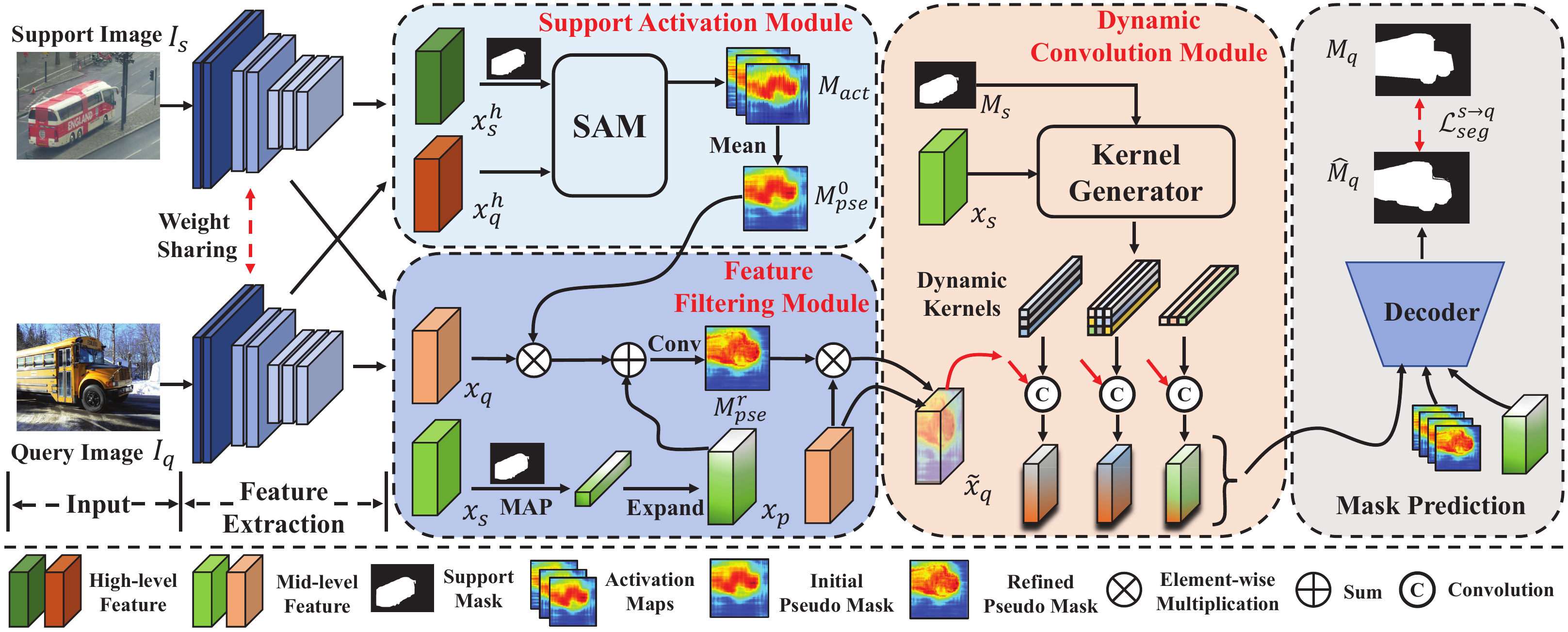}
	\caption{\textbf{Overall architecture of our proposed dynamic prototype convolution network (DPCN).} Firstly, \textbf{support activation module (SAM)} is introduced to generate activation maps and initial pseudo mask of target object in query image using high-level support and query feature. Then, \textbf{feature filtering module (FFM)} takes input mid-level support and query feature as well as initial pseudo mask to produce refined pseudo mask, which is leveraged to filter most background information in query feature. Meanwhile, \textbf{dynamic convolution module (DCM)} implements three groups of dynamic convolution over query features in parallel using kernels (multiple prototypes) generated from support foreground features, to propagate rich context information from support to query features. Finally, the updated features are concatenated and fed into a decoder for the final query segmentation mask prediction.
		\vspace{-4mm}}
	\label{fig2}
\end{figure*}

\subsection{Few-Shot Semantic Segmentation} 
Few-shot semantic segmentation (FSS) learns to segment target objects in query image given few pixel-wise annotated support images. Most existing FSS methods adopt two-branch architecture which 
implements meta-training on the base classes and then conducts meta-testing on the disjoint novel classes. OSLSM \cite{shaban2017one} is the first two-branch FSS model. Next, PL \cite{dong2018few} introduces prototype learning paradigm, which generates prototypes from support images to guide the segmentation of query objects. Recently, many prototype-based FSS methods emerge in the research
community, such as CANet \cite{Zhang_2019_CVPR}, SG-One \cite{zhang2020sg}, PANet \cite{wang2019panet}, PMMs \cite{yang2020prototype}, PFENet \cite{tian2020prior},and ASGNet \cite{Li_2021_CVPR}. The key idea of these methods lies in generating or rearranging representative prototypes using different strategies, then the interaction between prototypes with query features can be formulated as a few-to-many matching problem. However, these prototype learning methods inevitably cause information loss due to limited prototypes. Therefore, graph-based methods have thrived recently as they try to preserve structural information with many-to-many matching mechanism. For instance, PGNet \cite{Zhang_2019_CVPR} applies attentive graph reasoning to propagate label information from support data to query data. SAGNN \cite{xie2021scale} constructs graph nodes using multi-scale features and performs k-step reasoning over nodes to capture cross-scale information. Most recently, HSNet \cite{min2021hypercorrelation} proposes to tackle the FSS task from the perspective of visual correspondence. It implements efficient 4D convolutions over multi-level feature correlation and achieves great success. Different from previous methods, we try to perform sufficient interaction between support and query features using dynamic convolution, and mine as much complementary target information from both support and query features.

\subsection{Dynamic Convolution Networks}
Dynamic convolution networks aim to generate diverse kernels and implement convolution over input feature with these kernels. Many previous works have explored the effectiveness of dynamic convolution in deep neural networks.  DFN \cite{jia2016dynamic} proposes a dynamic filter network where filters are generated dynamically conditioned on input and achieves state-of-the-art performance on video and stereo prediction task. \cite{chen2020dynamic} aggregates multiple parallel convolution kernels dynamically based upon their attentions, and it boosts both image classification and keypoint detection accuracy. Dynamic convolution is also used in DMNet \cite{he2019dynamic} to adaptively capture multi-scale contents for predicting pixel-level semantic labels. The core of these methods is constructing multiple kernels from input features. Most recently, dynamic convolution is introduced into the few-shot object detection task by \cite{zhang2021accurate}, which generates various kernels from the object regions in support image and then  implements convolution over query feature using these  kernels, leading to a more representative query feature. In this paper, we propose to generate dynamic kernels from foreground support feature to interact with query feature by convolution. Instead of  only using square kennels as in \cite{zhang2021accurate}, we also introduce asymmetric kernels to capture subtle object details. Experiments in \textsection  \ref{ablation} demonstrate well the effectiveness of our method.

\section{Method}
\label{Methodology}
\subsection{Problem Setting}

We adopt the standard FSS setting, \emph{i.e.}, following the episode-based meta-learning paradigm \cite{snell2017prototypical}. We start from classes $\mathcal{C}_{tr}$ and $\mathcal{C}_{ts}$ for the training set $\mathcal{D}_{tr}$ and the test set $\mathcal{D}_{ts}$, respectively.
The key difference between FSS and general semantic segmentation task is that $\mathcal{C}_{tr}$ and $\mathcal{C}_{ts}$ in FSS are disjoint, $\mathcal{C}_{tr}\cap \mathcal{C}_{ts}=\emptyset$.
Both $\mathcal{D}_{tr}$ and $\mathcal{D}_{ts}$ consist of thousands of randomly sampled episodes, and each episode $(\mathcal{S}, \mathcal{Q})$ includes a support set $\mathcal{S}$, and a query set $\mathcal{Q}$ for a specific class $c$.
For the $k$-shot setting, the support set that contains $k$ image-mask pairs can be formulated as $\mathcal{S}=\{(I_s^{i}, M_s^{i})\}_{i=1}^k$, where $I_s^{i}$ represents $i$th support image and $M_s^{i}$ indicates corresponding binary mask. Similarly, we define the query set as $\mathcal{Q}=\{(I_q, M_q)\}$, where $I_q$ is the query image and its binary mask $M_q$ is only available in the model training phase.
In the meta-training stage, the FSS model  takes as input $\mathcal{S}$ and $I_q$ from a specific class $c$ and generates a predicted mask $\hat{M}_q$ for the query image.
Then the model can be trained with the supervision of a binary cross-entropy loss between $M_q$ and $\hat{M}_q$.
Finally, the model takes multiple randomly sampled episodes ${(S_i^{ts}, Q_i^{ts})}_{i=1}^{N_{ts}}$ from $\mathcal{D}_{ts}$ for evaluation. Next, the 1-shot setting is adopted to illustrate our method for simplicity.

\subsection{Overview}
\label{overview}

As in Fig.~\ref{fig2}, our dynamic prototype convolution network (DPCN) consists of three key modules, i.e., support activation module (SAM), feature filtering module (FFM), and dynamic convolution module (DCM). Specifically, given the support and query images, $I_s$ and $I_q$ , we use a common backbone with shared weights to extract both mid-level and high-level features.
We then have the SAM whose task is to generate an initial pseudo mask $M_{pse}^0$ for the target object in the query image. After SAM, a FFM follows, which aims to refine the pseudo mask and filter out irrelevant background information in the query feature. To incorporate relevant contextual information, we then employ the DCM, which learns to generate custom kernels from support foreground feature and employ dynamic convolution over query feature.
We then feed the pseudo masks and features computed by the dynamic convolutions into a decoder to predict the final segmentation mask $\hat{M}_q$ for the query image. 
Next, we describe each of the aforementioned modules in detail. 

\begin{figure}[!t]
	\centering
	\includegraphics[width=\columnwidth]{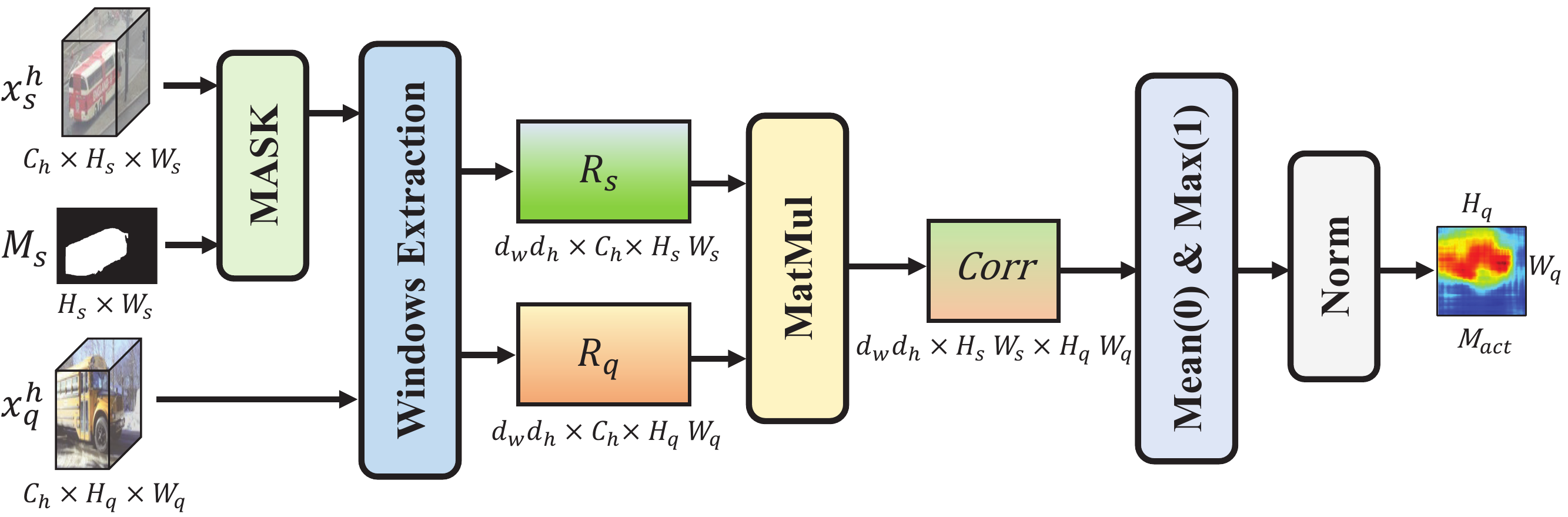} \vspace{-7mm}
	\caption{Illustration of the support activation module (SAM). }
	\label{fig3}
\end{figure}

\subsection{Support Activation Module}
\label{SAM}

Inspired by PFENet \cite{tian2020prior,luo2021pfenet}, recent FSS models \cite{xie2021scale, wu2021learning} usually leverage high-level features (e.g., \texttt{conv5} of ResNet50) from the support and query set to generate the prior mask indicating the rough location of the target object in the query image.
As this prior mask is usually obtained by element-to-element or square region-based matching between feature maps, a holistic context is not taken into account.

To counter this, with the support activation module we generate multiple activation maps of the target object in the query image using holistic region-to-region matching.
Specifically, as in Fig. \ref{fig3}, SAM takes as input the high-level support feature $x_s^h \in \mathbb{R}^{C_h \times H_s \times W_s}$, the corresponding binary mask $M_s \in \mathbb{R}^{H_s \times W_s}$, as well as the high-level query feature $x_q^h \in \mathbb{R}^{C_h \times H_q \times W_q}$, where $C_h$ is the channel dimension, $H_s, W_s, H_q, W_q$ are the height and width of support and query feature, respectively.

To perform holistic matching, we first need to generate region features $R_s$ and $R_q$ with a fixed window operation $\mathcal{W}$ sliding on the support and query features, respectively.
\begin{equation}
\begin{aligned}
& R_s=\mathcal{W}(x_s^h \otimes M_s) \in \mathbb{R}^{d_hd_w \times C_h \times H_sW_s}, \\
& R_q= \mathcal{W}(x_q^h) \in \mathbb{R}^{d_hd_w \times C_h \times H_qW_q},    
\end{aligned}    
\end{equation}
where $\otimes$ stands for the Hadamard product and $d_h, d_w$ are the window height and width.
In our experiments we opt for symmetrical and asymmetrical windows, i.e., $(d_h, d_w) \in \{(5,1), (3,3), (1,5)\}$ that are comprehensive and holistic regions, to account for possible object geometry variances.
Having the region features, we proceed with matching by computing their cosine similarity, which generates the regional matching map $Corr\in \mathbb{R}^{d_hd_w\times H_sW_s \times H_qW_q}$. 
 Notably, we utilize both square window (3,3) and asymmetrical windows (i.e., (5,1) and (1,5)), where square window can introduce more contextual information on regular part of target objects like the main body of object {\it plants}, asymmetrical windows can incorporate contextual details of slender part (e.g., leaves of {\it plants}).

We generate the final activation map $M_{act} \in \mathbb{R}^{H_q\times\ W_q}$ by taking the mean value among all regions and the maximal value among all support features followed by normalization operation.
As we have three windows, we have three activation maps, $\{M_{act}^i\}_{i=1}^3$.
In the end, we obtain the initial pseudo-mask $M_{pse}^0 \in \mathbb{R}^{H_q\times W_q}$, which indicates the rough location of target objects, by a mean operation.


\subsection{Feature Filtering Module}
\label{FFM}

As in Fig.~\ref{fig2}, the feature filtering module is constructed on mid-level support and query features, i.e.,
 $x_s\in \mathbb{R}^{C\times H \times W}$ and $x_q \in \mathbb{R}^{C\times H \times W}$ where $C, H, W$ are channel, height, and width, respectively. Given $x_s$, $x_q$, and the initial pseudo mask $M_{pse}^0$, the feature filtering module refines the pseudo mask, which is used to filter out irrelevant background information in the query image.
We first apply masked average pooling on the features from the support set to get prototype vector $p \in \mathbb{R}^{C\times 1 \times 1}$:
\begin{equation}
p = \text{average}(x_s \otimes \mathcal{R}(M_s)),
\end{equation}
where $\mathcal{R}$ reshapes support mask $M_s$ to be the same shape as $x_s$.
Then, we expand the support prototype vector $p$ to match the dimensions of the feature maps, $x_p \in \mathbb{R}^{C\times H \times W}$, and combine the target object information from both the support and query features.
We refine the pseudo mask with the help of a smaller network $\mathcal{F}$ composed of a 2D convolution layer followed by a sigmoid function,
\begin{equation}
M_{pse}^r = \mathcal{F}((x_q\otimes \mathcal{R}(M_{pse}^0)) \oplus x_p) \in \mathbb{R}^{H\times W},
\end{equation}
where $\oplus$ stands for the element-wise sum.
Compared with $M_{pse}^0$, $M_{pse}^r$ gives more accurate estimation of the object location in the query image.
Lastly, we obtain the final filtered query feature that discards irrelevant background by combining the feature $x_q$ with the prior mask :
\begin{equation}
\tilde{x}_q = (x_q \otimes M_{pse}^r) \oplus x_q \in \mathbb{R}^{C\times H\times W}.
\end{equation}
%

\subsection{Dynamic Convolution Module}
\label{DCM}
\begin{figure}[!t]
	\centering
	\includegraphics[width=\columnwidth]{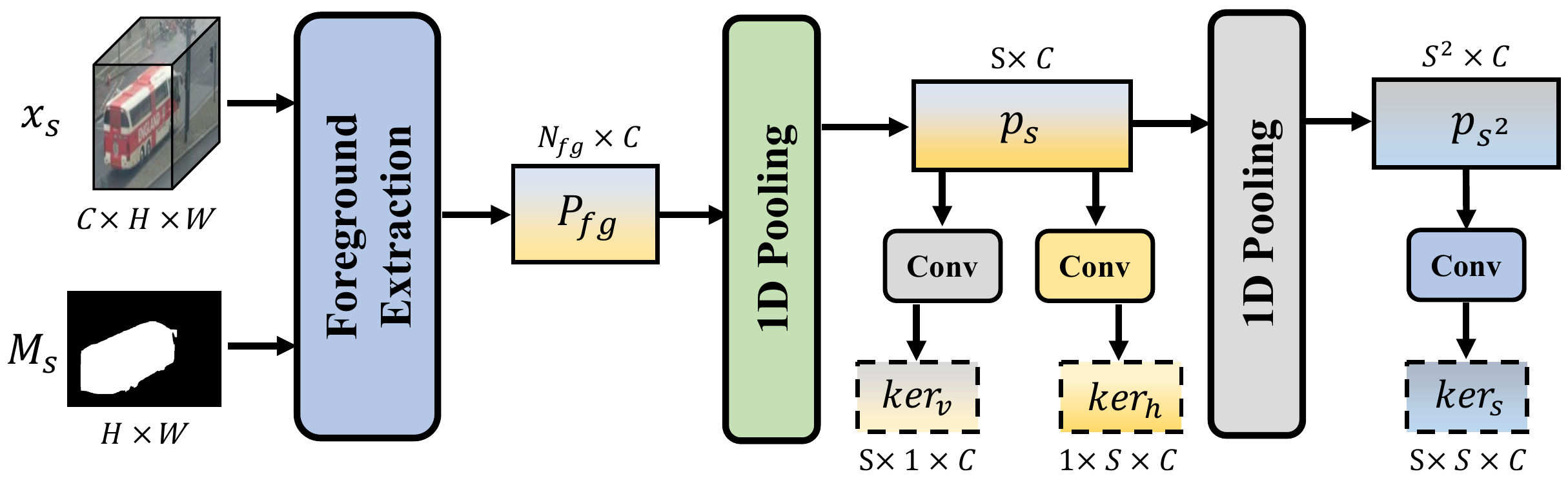}
	\caption{Illustration of kernel generator in DCM. \vspace{-5mm}}
	\label{fig4}
\end{figure}
In the previous step we obtain a foreground feature from the query, which is minimally affected by irrelevant background.
Still, the operations so far have been so that to provide a rough estimate of the location of the target object.
For accurate segmentation, however, much finer pixel-level predictions are required.
In the absence of significant data to train our filters on, we introduce dynamic convolutions. We illustrate DCM in Fig. \ref{fig2}, and Fig. \ref{fig4} depicts the details of the kernel generator.

Dynamic convolutions rely on meta-learning to infer what are the optimal kernel parameters given a subset of features, agnostic to the unknown underlying class semantics.
Specifically, we input the mid-level support feature $x_s$ and the corresponding mask $M_s$ to a kernel generator, which generates dynamic kernels, i.e., one group square kernel and two groups of asymmetrical kernels.
Then, we carry out three convolution operations over the filtered query feature $\tilde{x}_q$ using dynamic kernels. 
Firstly, we extract foreground vectors $P_{fg}$ from support feature:
\begin{equation}
\label{eq7}
P_{fg} = \mathcal{F}_e(x_s\otimes M_s) \in \mathbb{R}^{N_{fg}\times C},
\end{equation}
where $\mathcal{F}_e$ is the foreground extraction function without any learnable parameters, $N_{fg}$ represents the number of foreground vectors.
Next, two consecutive 1D pooling operations with kernel size $S$ and $S^2$ are leveraged to obtain two groups of prototypes $p_s \in \mathbb{R}^{S \times C}$ and $p_{s^2} \in \mathbb{R}^{S^2 \times C}$:
\begin{equation}
p_s = pool_s(P_{fg}),  p_{s^2}=pool_{s^2}(p_s).
\end{equation}
As discussed above, we achieve dynamic convolution over query feature using a square kernel and two asymmetric kernels.
As such, we use three parallel convolutional neural networks whose outputs are the generated kernel weights:
\begin{equation}
\begin{aligned}
& ker_{v} = \mathcal{F}_{conv1}(p_s) \in \mathbb{R}^{S\times 1\times C},\\
& ker_{h} = \mathcal{F}_{conv2}(p_s) \in \mathbb{R}^{1\times S\times C},\\
& ker_{s} = \mathcal{F}_{conv3}(p_{s^2}) \in \mathbb{R}^{S\times S\times C},
\end{aligned}
\end{equation}
where $ker_v, ker_h, ker_s$ are the vertical, horizontal, and square kernel weights, respectively.
$\mathcal{F}_{conv1}$, $\mathcal{F}_{conv2}$, and $\mathcal{F}_{conv3}$ represent corresponding convolution sub networks, which are achieved by two consecutive 1D convolution layers. We emphasize that the above parameter generating networks \emph{do not} share parameters.
Given the vertical kernel $ker_v$, the query feature $\tilde{x}_q$ can be enhanced as $\tilde{x}_q^v \in \mathbb{R}^{C\times H\times W}$:
\begin{equation}
\tilde{x}_q^v =  \mathcal{F}_{dc}(\tilde{x}_q|ker_v),
\end{equation}
where $\mathcal{F}_{dc}$ denotes the dynamic convolution operation, and $ker_v$ works as the kernel weight. Similarly, we can obtain other enhanced query features $\tilde{x}_q^h \in \mathbb{R}^{C\times H\times W}$ and $\tilde{x}_q^s \in \mathbb{R}^{C\times H\times W}$ with horizontal kernel $ker_h$ and square kernel $ker_s$, respectively.
With the sufficient interaction between query feature and dynamic support kernels, the object context in the generated query feature are enhanced.

Then, the enhanced query features ${\tilde{x}_q^v, \tilde{x}_q^h, \tilde{x}_q^s}$, support foreground feature $x_p$, support activation maps $\{M_{act}^i\}_{i=1}^3$, and refined pseudo mask $M_{pse}^r$ are all reshaped to the same spatial size and concatenated to a representative feature $x_{out} \in \mathbb{R}^{(4C+4)\times H \times W} $:
\begin{equation}
x_{out} = \mathcal{F}_{cat}(\tilde{x}_q^v, \tilde{x}_q^h, \tilde{x}_q^s,x_p,\{M_{act}^i\}_{i=1}^3, M_{pse}^r),
\end{equation}
where $\mathcal{F}_{cat}$ is the concatenation operation in channel dimension. Finally, $x_{out}$ is fed into a decoder to generate segmentation mask $\hat{M}_q$ for query image $I_q$:
\begin{equation}
\label{eq12}
\hat{M}_q = \mathcal{F}_{cls}(\mathcal{F}_{ASPP}(\mathcal{F}_{conv}(x_{out}))),
\end{equation}
where $\mathcal{F}_{conv}$, $\mathcal{F}_{ASPP}$, and $\mathcal{F}_{cls}$ are three consecutive modules that constitute the decoder.

\subsection{Extension to $k$-shot setting}
\label{k-shot}

So far we have focused on the one-shot setting, summarized in Fig. \ref{fig2}.
For the $k$-shot setting, where more than one support images are available, most existing methods choose attention-based fusion or feature averaging.
However, such a simple strategy does not make full use of the support information.
By contrast, we can easily extend the dynamic convolutions to the $k$-shot setting and achieve substantial performance improvement.
Specifically, given each support image-mask pair, we extract foreground vectors with Eq. (\ref{eq7}).
By collecting all foreground vectors together, we get the overall support foreground vectors $P_{fg}$ from $k$ shots:
\begin{equation}
P_{fg} = (P_{fg}^1, P_{fg}^2,\dots, P_{fg}^k) \in \mathbb{R}^{N_{fg}\times C},
\end{equation}
where the number of foreground vectors is $N_{fg}=\sum_{i=1}^kN_{fg}^i$.
By doing so, the kernel generator in DCM can generate more robust dynamic kernels, thus leading to more adequate interaction and accurate query mask estimation.

\subsection{Training Loss}
\label{loss}

Our dynamic prototype convolution network is trained in an end-to-end manner with the binary cross-entropy loss (BCE).
Given predicted mask $\hat{M}_q$ and ground-truth mask $M_q$ for query image $I_q$, we formulate the BCE loss between $\hat{M}_q$ and $M_q$ as our main loss:
\begin{equation}
\mathcal{L}_{seg}^{s\rightarrow q}= \frac{1}{hw}\sum_{i=1}^h\sum_{j=1}^wBCE(\hat{M}_q(i,j), {M}_q(i, j)).
\end{equation}

Inspired by \cite{wang2019panet}, we implement another branch to estimate the support mask using query image $I_q$ and its corresponding predicted mask $\hat{M}_q$.
Similar with Eq. \eqref{eq12}, we get the predicted support mask $\hat{M}_s$.
Then we get another loss by calculating BCE loss between $\hat{M}_s$ and $M_s$:
\begin{equation}
\mathcal{L}_{seg}^{q\rightarrow s}= \frac{1}{h_sw_s}\sum_{i=1}^{h_s}\sum_{j=1}^{w_s}BCE(\hat{M}_s(i,j), {M}_s(i, j)),
\end{equation}
where $h_s$ and $w_s$ are the height and width of ground-truth mask $M_s$ for support image $I_s$. 
Note that both query and support mask prediction process share the same structure and parameters.
In summary, the final loss is:
\begin{equation}
\label{lambda}
\mathcal{L} = \mathcal{L}_{seg}^{s\rightarrow q} + \lambda \mathcal{L}_{seg}^{q\rightarrow s}, 
\end{equation}
where $\lambda$ is weight to balance the contribution of each branch and set to 1.0 in all experiments.

\Crefname{table}{Table}{Tables}
\crefname{table}{Tab.}{Tabs.}

\section{Experiments}
\label{Experiments}



\subsection{Experimental Settings}

\textbf{Datasets.}
We follow \cite{tian2020prior} and adopt PASCAL-$5^i$ \cite{shaban2017one} and COCO-$20^i$ \cite{nguyen2019feature} benchmarks for evaluation.
PASCAL-$5^i$ comes from PASCAL VOC2012 \cite{everingham2010pascal} and additional SBD \cite{hariharan2011semantic} annotations.
It contains 20 object classes split into 4 folds, which are used for 4-fold cross validation.
For each fold, 5 classes are used for testing and the remaining 15 classes for training. COCO-$20^i$ is a more challenging benchmark, which is created from MSCOCO \cite{lin2014microsoft} and contains 80 object classes.
Similarly, we split classes in COCO-$20^i$ into 4 folds with 20 classes per fold,  for each fold, we utilize 20 classes for testing and the remaining 60 classes for training.

\textbf{Metrics and Evaluation.}
Following \cite{tian2020prior,xie2021scale,min2021hypercorrelation}, mean intersection over union (mIoU) and foreground-background IoU (FB-IoU) are adopted as our metrics for evaluation.
While FB-IoU neglects object classes and directly averages foreground and background IoU, mIoU averages IoU values of all classes in a fold. During evaluation, 1,000 episodes are sampled from the test set for metrics calculation, and multi-scale testing is used like most existing FSS methods.

\begin{table*}[]
	\centering
	\renewcommand\arraystretch{1}
	\resizebox{\linewidth}{!}{
		\begin{tabular}{r|c|cccccc|cccccc}
			\toprule[2pt]
			
			\multirow{2}{*}{Methods} & \multicolumn{1}{c|}{\multirow{2}{*}{Backbone}} & \multicolumn{6}{c|}{1-shot}  & \multicolumn{6}{c}{5-shot}  \\
			& \multicolumn{1}{c|}{} 
			& \multicolumn{1}{l}{Fold-0} & Fold-1  & Fold-2  & Fold-3   & Mean  & FB-IoU
			& \multicolumn{1}{l}{Fold-0} & Fold-1  & Fold-2  & Fold-3   & Mean  & FB-IoU \\ \hline
			OSLSM (BMVC'17) \cite{shaban2017one} & VGG16                         
			& 33.6  & 55.3  & 40.9  & 33.5   & 40.8  & 61.3                 
			& 35.9  & 58.1  & 42.7  & 39.1   & 43.9  & 61.5           \\
			co-FCN (ICLRW'18) \cite{rakelly2018conditional}  & VGG16
			& 36.7  & 50.6  & 44.9  & 32.4   & 41.1  & 60.1   
			& 37.5  & 50.0  & 44.1  & 33.9   & 41.4  & 60.2            \\
			AMP-2(ICCV'19) \cite{siam2019amp}   & VGG16
			& 41.9  & 50.2  & 46.7  & 34.7   & 43.4  & 61.9 
			& 40.3  & 55.3  & 49.9  & 40.1   & 46.4  & 62.1            \\
			PFENet (TPAMI’20) \cite{tian2020prior}  &  VGG16
			& 56.9  & 68.2  & 54.4  & 52.4   & 58.0  & 72.0  
			& 59.0  & 69.1  & 54.8  & 52.9   & 59.0  & 72.3            \\
			HSNet (ICCV'21) \cite{min2021hypercorrelation}  & VGG16
			& 59.6  & 65.7  & 59.6  & 54.0   & 59.7  & 73.4 
			& 64.9  & 69.0  & 64.1  & 58.6   & 64.1  & 76.6            \\ \hline
			PFENet (TPAMI’20) \cite{tian2020prior}  & ResNet50
			& 61.7  & 69.5  & 55.4  & 56.3   & 60.8  & 73.3  
			& 63.1  & 70.7  & 55.8  & 57.9   & 61.9  & 73.9            \\
			RePRI (CVPR'21) \cite{boudiaf2021few}   &  ResNet50
			& 59.8  & 68.3  & 62.1  & 48.5   & 59.7  & -    
			& 64.6  & 71.4  & \textbf{71.1}  & 59.3   & 66.6  & -               \\
			SAGNN (CVPR'21) \cite{xie2021scale}  & ResNet50
			& 64.7  & 69.6  & 57.0  & 57.3   & 62.1  & 73.2 
			& 64.9  & 70.0  & 57.0  & 59.3   & 62.8  & 73.3            \\
			SCL (CVPR'21) \cite{zhang2021self}  & ResNet50
			& 63.0  & 70.0  & 56.5  & 57.7   & 61.8  & 71.9 
			& 64.5  & 70.9  & 57.3  & 58.7   & 62.9  & 72.8            \\
			MLC (ICCV'21) \cite{yang2021mining}  & ResNet50
			& 59.2  & 71.2  & 65.6  & 52.5   & 62.1  & - 
			& 63.5  & 71.6  & 71.2  & 58.1   & 66.1  & -               \\
			MMNet (ICCV'21) \cite{wu2021learning}  & ResNet50
			& 62.7  & 70.2  & 57.3  & 57.0   & 61.8  & -
			& 62.2  & 71.5  & 57.5  & 62.4   & 63.4  & -               \\
			HSNet (ICCV'21) \cite{min2021hypercorrelation}  & ResNet50
			& 64.3  & 70.7  & 60.3  & 60.5   & 64.0  & 76.7  
			& \textbf{70.3}  & \textbf{73.2}  & 67.4  & \textbf{67.1}   & 69.5  & 80.6             \\ \hline
			\textbf{Baseline}    & VGG16
			& 58.4  & 68.0  & 58.0  & 50.9  & 58.8  & 71.2  
			& 60.7  & 68.8  & 60.2  & 52.2  & 60.4  & 74.3                                     \\
			\textbf{DPCN} & VGG16 
			&  58.9 & 69.1  & 63.2  & 55.7  & 61.7  &  73.7 
			&  63.4 & 70.7  & 68.1  & 59.0  & 65.3  & 77.2             \\ 
			\textbf{Baseline}    & ResNet50
			& 61.1  & 69.8  & 58.4   & 56.3   & 61.4   & 71.5   
			& 63.7  & 70.9  & 58.7  & 57.4  & 62.7  & 73.7                                      \\
			\textbf{DPCN} & ResNet50
			& \textbf{65.7}  & \textbf{71.6}  & \textbf{69.1}  & \textbf{60.6}  & \textbf{66.7}  & \textbf{78.0}  
			& 70.0  & \textbf{73.2}  & 70.9  & 65.5  & \textbf{69.9}  & \textbf{80.7}               \\ 
			
			\bottomrule[2pt]
		\end{tabular}}
		\vspace{-2mm}
		\caption{Comparison with state-of-the-arts on PASCAL-$5^i$ dataset under both 1-shot and 5-shot settings. mIoU of each fold, and averaged mIoU \& FB-IoU of all folds are reported. Baseline results are achieved by removing three modules (i.e., SAM, FFM, and DCM) in DPCN.\vspace{-2mm}}
		\label{pascal}
\end{table*}
	
\begin{table*}[]
	\centering
	\centering
	\renewcommand\arraystretch{1}
	\resizebox{\linewidth}{!}{
		\begin{tabular}{r|c|cccccc|cccccc}
			\toprule[2pt]
			
			\multirow{2}{*}{Methods} & \multicolumn{1}{c|}{\multirow{2}{*}{Backbone}} & \multicolumn{6}{c|}{1-shot}  & \multicolumn{6}{c}{5-shot}  \\
			& \multicolumn{1}{c|}{} 
			& \multicolumn{1}{l}{Fold-0} & Fold-1  & Fold-2  & Fold-3   & Mean  & FB-IoU
			& \multicolumn{1}{l}{Fold-0} & Fold-1  & Fold-2  & Fold-3   & Mean  & FB-IoU \\ \hline
			FWB(ICCV'19) \cite{nguyen2019feature} & VGG16    
			& 18.4   & 16.7   & 19.6   & 25.4   & 20.0 & -     
			& 20.9   & 19.2   & 21.9   & 28.4   & 22.6 & -    \\
			PFENet(TPAMI’20) \cite{tian2020prior} & VGG16
			& 33.4  & 36.0    & 34.1   & 32.8   & 34.1 & 60.0   
			& 35.9  & 40.7   & 38.1   & 36.1   & 37.7 & 61.6    \\
			SAGNN(CVPR'21) \cite{xie2021scale} & VGG16  
			& 35.0  & 40.5   & 37.6   & 36.0   & 37.3 & 61.2  
			& 37.2  & 45.2   & 40.4   & 40.0   & 40.7 & 63.1  \\ 
			\hline
			RePRI(CVPR'21) \cite{boudiaf2021few} & ResNet50  
			& 31.2  & 38.1   & 33.3   & 33.0   & 34.0 & -      
			& 38.5  & 46.2   & 40.0   & 43.6   & 42.1 & -    \\
			MLC(ICCV'21) \cite{yang2021mining}  & ResNet50
			& \textbf{46.8}  & 35.3   & 26.2   & 27.1   & 33.9 & -      
			& \textbf{54.1}  & 41.2   & 34.1   & 33.1   & 40.6 & -     \\
			MMNet(ICCV'21) \cite{wu2021learning}  & ResNet50 
			& 34.9  & 41.0   & 37.2   & 37.0   & 37.5 & -  
			& 37.0  & 40.3   & 39.3   & 36.0   & 38.2 & -     \\
			HSNet(ICCV'21) \cite{min2021hypercorrelation} & ResNet50  
			& 36.3  & 43.1   & 38.7   & 38.7   & 39.2 & \textbf{68.2}  
			& 43.3  & 51.3   & 48.2   & 45.0   & 46.9 & \textbf{70.7}   \\
			SAGNN(CVPR'21) \cite{xie2021scale} & ResNet101 
			& 36.1  & 41.0   & 38.2   & 33.5   & 37.2 & 60.9  
			& 40.9  & 48.3   & 42.6   & 38.9   & 42.7 & 63.4    \\ 
			SCL(CVPR'21) \cite{zhang2021self}  & ResNet101  
			& 36.4  & 38.6   & 37.5   & 35.4   & 37.0 & - 
			& 38.9  & 40.5   & 41.5   & 38.7   & 39.9 & -       \\ \hline
			\textbf{Baseline} & VGG16                    
			& 32.1   & 36.1   & 35.2   & 32.3   & 33.9    & 60.1  
			& 35.0   & 40.1    & 37.1   & 36.5  & 37.2    &  61.8       \\
			\textbf{DPCN} & VGG16 
			& 38.5  & 43.7   & 38.2  & 37.7  & 39.5   & 62.5 
			& 42.7  & 51.6   & 45.7  & 44.6  & 46.2   & 66.1                               \\ 
			\textbf{Baseline} & ResNet50                    
			& 32.3  & 38.3   & 34.9  & 32.5  & 34.5   & 57.7 
			& 35.0  & 41.0   & 37.3  & 35.5  & 37.2   & 59.2        \\
			\textbf{DPCN} & ResNet50 
			& 42.0  & \textbf{47.0}   & \textbf{43.2}  & \textbf{39.7}  & \textbf{43.0}   & 63.2 
			& 46.0  & \textbf{54.9}   & \textbf{50.8}  & \textbf{47.4}  & \textbf{49.8}  & 67.4                               \\ 
			
			\bottomrule[2pt]
		\end{tabular}}
		\vspace{-2mm}
		\caption{Comparison with state-of-the-arts on COCO-$20^i$ dataset under both 1-shot and 5-shot settings. mIoU of each fold, and averaged mIoU \& FB-IoU of all folds are reported. Baseline results are achieved by removing three modules (i.e., SAM, FFM, and DCM) in DPCN. 
			\vspace{-7mm}}
		\label{coco}
\end{table*}

\textbf{Implementation Details.}
We use ResNet-50 \cite{he2016deep} and VGG-16 \cite{simonyan2014very} pre-trained on ImageNet as our backbone networks.
The backbone weights are fixed except for layer4, which is required to learn more robust activation maps on PASCAL-$5^i$, meanwhile, all these weights are fixed for COCO-$20^i$ to pursue a faster training time.
The model is trained with the SGD optimizer on PASCAL-$5^i$ for 200 epochs and COCO-$20^i$ for 50 epochs.
The learning rate is initialized as 0.005 with batch size 8 (0.002 with batch size 32) for PASCAL-$5^i$ (COCO-$20^i$).
Data augmentation strategies in \cite{tian2020prior} are adopted in the training stage, and all images are cropped to $473\times473$ patches for two benchmarks.
In addition, the window sizes in SAM are set to $\{5\times1, 3\times3, 1\times5\}$ for PASCAL-$5^i$; since COCO-$20^i$ contains much larger amounts of training images with plentiful types (already having holistic contexts compared with images in PASCAL-$5^i$), we only utilize the initial prior mask for speeding up our model training. The kernel sizes in DCM are set as $5\times1$, $5\times5$, and $1\times5$ for two datasets.
We implement our model with PyTorch1.7 and conduct all the experiments with Nvidia Tesla V100 GPUs.

\subsection{Comparisons with State-of-the-Arts}
\textbf{PASCAL-$5^i$.}
We report the mIoU and FB-IoU under both 1-shot and 5-shot settings in Table~\ref{pascal}.
It can be seen that \textbf{(i)} DPCN achieves state-of-the-art performance under both 1-shot and 5-shot settings.
Especially for the 1-shot setting, we surpass HSNet \cite{min2021hypercorrelation}, which holds the previous state-of-the-art results, by 2.0\% and 2.7\% with VGG16 and ResNet50 as backbone networks, respectively.
In addition, DPCN also presents comparable performance with HSNet under 5-shot setting while using less mid-level features.
\textbf{(ii)} DPCN outperforms its baseline method with a large margin (e.g., mIoU 66.7\% versus 61.4\% with ResNet50 backbone for 1-shot setting), which is implemented with same architecture except for proposed components (i.e., SAM, FFM, and DCM).
The results further demonstrate that DPCN can effectively mine complementary information from both support and query features to facilitate query image segmentation.
\begin{figure*}[!t]
	\centering
	\includegraphics[width=\textwidth]{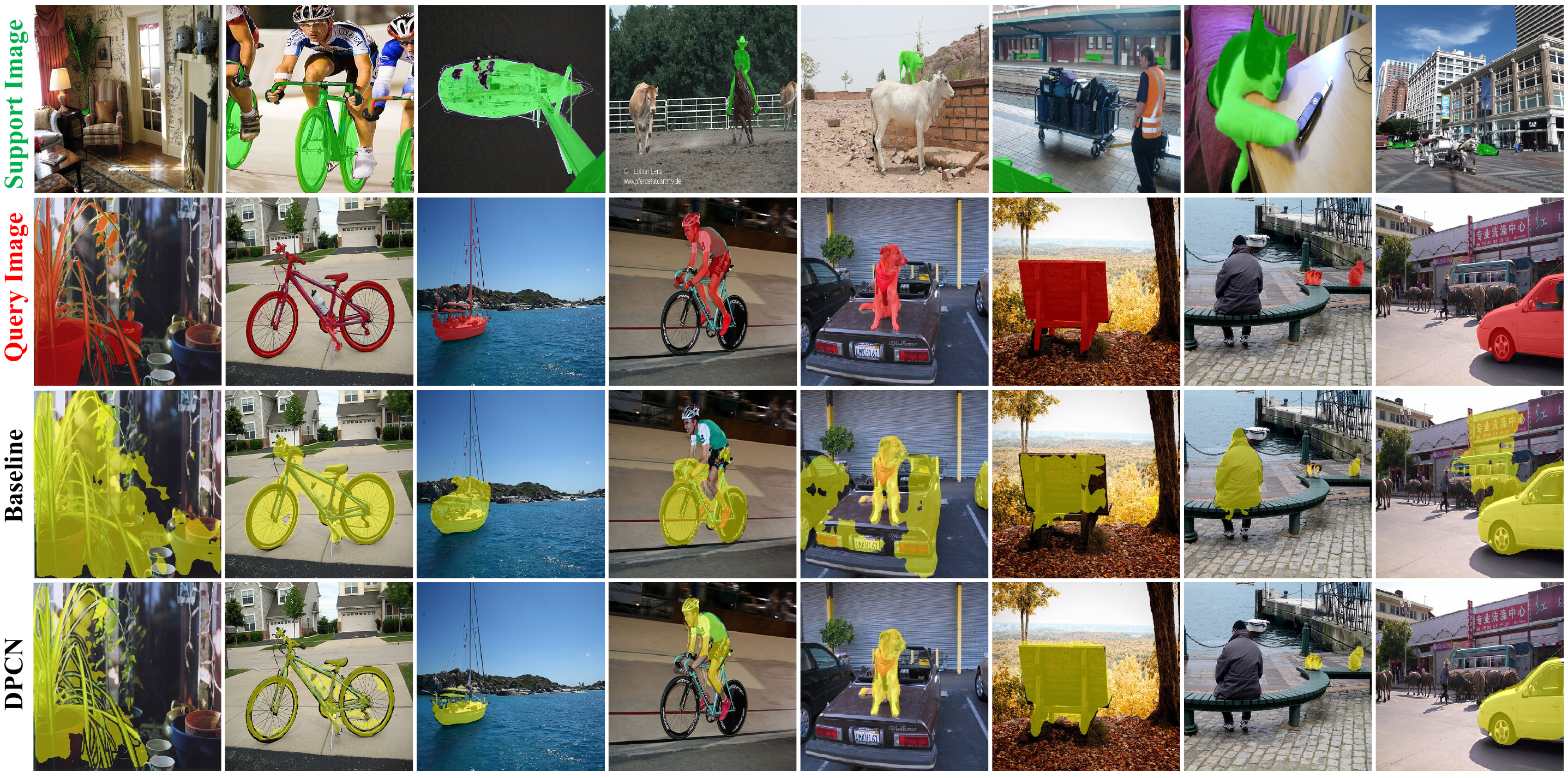}
	\vspace{-8mm}
	\caption{Qualitative results of our method DPCN and baseline model on PASCAL-$5^i$ and COCO-$20^i$ benchmarks. Zoom in for details.\vspace{-4mm}}
	\label{fig5}
\end{figure*}

\textbf{COCO-$20^i$.}
COCO-$20^i$ is a more challenging benchmark which usually contains multiple objects and exhibits great variance.
Table \ref{coco} presents the performance comparison of mIoU and FB-IoU on COCO-$20^i$ dataset.
As can be seen, using VGG16 and ResNet50 as backbone, our model DPCN significantly outperforms recent methods under both 1-shot and 5-shot settings.
With the ResNet50 backbone, DPCN achieves 3.8\% and 2.9\% of mIoU improvement over HSNet \cite{min2021hypercorrelation} (previous SOTA) under 1-shot and 5-shot settings, respectively.
In addition, DPCN gains significant improvement over the baseline models.
For example, DPCN with VGG16 backbone achieves 5.6\%  and 9.0\% mIoU improvement over the baseline model, which proves the superiority of our model in such challenging scenarios.

\begin{table}[!t]
	\centering
	\renewcommand\arraystretch{1}
	\resizebox{\columnwidth}{!}{
		\begin{tabular}{ccc|ccccc|c}
			\toprule[2pt]
			
			&  &   & \multicolumn{5}{c|}{1-shot mIoU} &  \\ 
			\multirow{-2}{*}{SAM} & \multirow{-2}{*}{FFM} & \multirow{-2}{*}{DCM} & \multicolumn{1}{l}{Fold-0} & \multicolumn{1}{l}{Fold-1} & \multicolumn{1}{l}{Fold-2} & \multicolumn{1}{l}{Fold-3} & \multicolumn{1}{l|}{Mean}
			& \multirow{-2}{*}{FB-IoU} \\ \hline
			\Checkmark & \Checkmark  &     & 63.6   & 69.7  & 65.0  & 59.6  & 64.5  & 75.2   \\
			&\Checkmark  & \Checkmark & \textbf{67.1} & 71.1& 63.2  & 60.0 & 65.4 & 76.0  \\
			\Checkmark  &  & \Checkmark   & 63.5  & 70.8   & 65.7  & 58.9  & 64.7   & 75.8 \\
			\Checkmark & \Checkmark  & \Checkmark  & 65.7  & \textbf{71.6}  & \textbf{69.1}  & \textbf{60.6}   & \textbf{66.7}  & \textbf{78.0}  \\ 
			
			\bottomrule[2pt]
		\end{tabular}}
		\vspace{-4mm}
		\caption{Ablation studies of main model components.
			\vspace{-6mm}}
		\label{ablation1}
	\end{table}
\begin{table}[!t]
	\centering
	\renewcommand\arraystretch{1}
	\resizebox{\columnwidth}{!}{
		\begin{tabular}{c|ccccc|c}
			\toprule[2pt]
			
			& \multicolumn{5}{c|}{1-shot mIoU} & \\ 
			\multirow{-2}{*}{Kernel Size} & Fold-0 & Fold-1 & Fold-2 & Fold-3 & Mean    
			& \multirow{-2}{*}{FB-IoU} \\ \hline
			3   & 65.2   & 70.4   & 68.5  & 59.4  & 65.9   & 77.5  \\
			5   & 65.7   & \textbf{71.6}   & 69.1  & \textbf{60.6}  & \textbf{66.7}   & \textbf{78.0}   \\
			7   & 65.5   & 70.7   & \textbf{69.3}  & 59.0  & 66.1   & 77.5  \\
			9   & \textbf{65.9}   & 70.8   & 68.8  & 59.7  & 66.3   & 77.7  \\
			
			\bottomrule[2pt]
		\end{tabular}}
		\vspace{-4mm}
		\caption{Ablation studies on kernel size.
		\vspace{-6mm}}
		\label{ablation3}
	\end{table}
\textbf{Qualitative Results.}
In Fig. \ref{fig5}, we report some quantitative results generated from our DPCN and baseline model on the PASCAL-$5^i$ and COCO-$20^i$ benchmarks.
Compared with the baseline, we can see that DPCN exhibits better performance in capturing object details.
For instance, more tiny details are preserved in the segmentation of {\it plants} and {\it bike} (first two columns in Fig. \ref{fig5}).
Refer to supplementary materials for more qualitative results.

\subsection{Ablations}
\label{ablation}
We conduct following ablation studies with ResNet-50 backbone under the 1-shot setting on PASCAL-$5^i$ dataset.

\textbf{Components Analysis.}
DPCN contains three major components, i.e., support activation module (SAM), feature filtering module (FFM), and dynamic convolution module (DCM). Table \ref{ablation1} presents our validation on the effectiveness of each component. DCM, which is the most important component in our model, brings 2.2\% improvement in mIoU. Meantime, SAM and FFM are also indispensable. By combing all three modules, DPCN achieves state-of-the-art performance.

\textbf{Kernel Size in DCM.}
We take the kernel size from $\{3,5,7,9\}$ to investigate the performance of our DPCN model. We can see from Table \ref{ablation3} that DPCN achieves the best and second-best performance when kernel sizes are 5 and 9, respectively. By the way, the performance drops slightly when the kernel sizes are 3 and 7. Therefore, the kernel size is set as 5 in all of our experiments.

\begin{table}[!t]
	\centering
	\renewcommand\arraystretch{1}
	\resizebox{\columnwidth}{!}{
		\begin{tabular}{c|ccccc|c}
			\toprule[2pt]
			
			& \multicolumn{5}{c|}{1-shot mIoU} &       \\ 
			\multirow{-2}{*}{Kernel variants} & \multicolumn{1}{l}{Fold-0} & \multicolumn{1}{l}{Fold-1} & \multicolumn{1}{l}{Fold-2}  & \multicolumn{1}{l}{Fold-3} & \multicolumn{1}{l|}{Mean}   & \multirow{-2}{*}{FB-IoU} \\ \hline
			w/o DCM        & 63.6  & 69.7  & 65.0 & 59.6  & 64.5  & 75.2  \\
			$5\times5$     & 64.7  & 71.2  & 65.3 & 58.7  & 65.0  & 76.2      \\
			$1\times5$     & 64.9  & 71.0  & 65.2 & 59.7  & 65.2  & 75.4  \\
			$1\times5, 5\times1 $  & 63.2 & 71.4 & 64.8 & 59.2 & 64.7 & 75.4  \\
			$1\times5, 5\times5, 5\times1 $ & \textbf{65.7} & \textbf{71.6} & \textbf{69.1} & \textbf{60.6} & \textbf{66.7} 
			& \textbf{78.0} \\ 
			
			\bottomrule[2pt]
		\end{tabular}}
		\vspace{-2mm}
		\caption{Ablation studies on different kernel variants of DCM.}
		\label{ablation2}
		\vspace{-3mm}
	\end{table}
	\begin{table}[!t]
		\centering
		\renewcommand\arraystretch{1.1}
		\resizebox{\columnwidth}{!}{
			\begin{tabular}{c|ccccc|c}
				\toprule[2pt]
				
				& \multicolumn{5}{c|}{1-shot mIoU} & \\ 
				\multirow{-2}{*}{Methods} & Fold-0 & Fold-1 & Fold-2 & Fold-3 & Mean    
				& \multirow{-2}{*}{FB-IoU} \\ \hline
				CANet       & 53.5   & 65.9   & 51.3  & 51.9  & 55.4   & 66.2  \\
				CANet+DCM   & 64.7   & 66.8   & 51.8  & 51.9  & 58.8   &  69.3    \\
				PFENet      & 61.7   & 69.5   & 55.4  & 56.3  & 60.8 & 73.3  \\
				PFENet+DCM  & 62.2   & 69.6   & 59.2  & 58.0  & 62.3   & 73.5  \\
				
				\bottomrule[2pt]
			\end{tabular}}
			\vspace{-3mm}
			\caption{Generalization ability of the proposed DCM.
			\vspace{-5mm}}
			\label{ablation4}
		\end{table}		
\textbf{Kernel Variants in DCM.}
Dynamic kernels are important components in DCM, so we evaluate the effectiveness of different kernel variants. As shown in Table \ref{ablation2}, square kernel and asymmetric kernels achieve almost similar results. However, DPCN yields better performance when choosing both square kernel and asymmetric kernels as dynamic kernels.

\textbf{Generalization of DCM.}
DCM can be utilized as a plug-and-play module to further improve current prototype-based methods. To verify this, we apply DCM to CANet \cite{Zhang_2019_CVPR} and PFENet \cite{tian2020prior}. As shown in Table \ref{ablation4}, DCM brings significant improvements on both CANet and PFENet.

\textbf{5-shot Fusion Strategies.}
We compare our 5-shot fusion strategy (discussed in \textsection  \ref{k-shot} ) with voting strategy \cite{min2021hypercorrelation}, average \cite{shaban2017one} and OR \cite{Zhang_2019_CVPR} strategies on masks in Table \ref{ablation5}. We can see that our 5-shot fusion strategy achieves 3.2\% mIoU improvement and outperforms other fusion strategies.

\section{Conclusion}
\vspace{-2mm}
\label{Conclusion}
We propose a dynamic prototype convolution network (DPCN) with three major components (i.e., SAM, FFM, and DCM) to address the challenging FSS task. To better mine information from query image, we propose SAM and FFM to generate pseudo query mask and filter background information, respectively. Moreover, a plug-and-play module DCM is designed to implement sufficient interaction between support and query features. Extensive experiments demonstrate that DPCN achieves state-of-the-art results. 
\section*{Acknowledgement}
This work was partially funded by Elekta Oncology Systems AB and a RVO public-private partnership grant (PPS2102).

\clearpage

{\small
\bibliographystyle{ieee_fullname}
\bibliography{Main}
}

\renewcommand{\thetable}{A\arabic{table}}
\renewcommand{\thefigure}{A\arabic{figure}}
\section*{Appendix A. Implementation details}
\renewcommand{\thetable}{A\arabic{table}}
We employ ResNet-50 \cite{chen2020dynamic} (VGG-16 \cite{simonyan2014very}) pre-trained on ImageNet as our backbone networks.
For ResNet-50, the dilation convolution (with stride size = 1) is introduced to ensure that the feature receptive fields of layer2, layer3, and layer4 preserve the same spatial resolution. 
The backbone weights are fixed except for layer4, which is required to learn more robust activation maps.
The model is trained with a SGD optimizer for 200 and 50 epochs on the PASCAL-$5^i$ and the COCO-$20^i$ benchmarks, respectively .
The learning rates are initialized as 0.005 and 0.002 with a poly learning rate schedule on PASCAL-$5^i$ and COCO-$20^i$, respectively. The batch size is set as 8 on PASCAL-$5^i$ and 32 on COCO-$20^i$. Our entire network is trained with the same learning rate during each epoch, except for layer4 of the backbone network, whose parameters starts back-propagation after training for multiple epochs to ensure a lower learning rate for fine-tuning.
Data augmentation strategies in \cite{tian2020prior} are adopted in the training stage, and all images are cropped to $473\times473$ patches for two benchmarks. Besides, we leverage multi-scale testing strategy used in the most few-shot semantic segmentation methods for the model evaluation, and the original Groudtruth of the evaluated query image without any resize operations is  adopted for the metric calculation.
In addition, the window sizes in SAM are set to $\{5\times1, 3\times3, 1\times5\}$, and the kernel sizes in DCM are set as $5\times1$, $5\times5$, and $1\times5$, respectively.
We implement our model with PyTorch 1.7.0 and conduct all the experiments with Nvidia Tesla V100 GPUs and CUDA11.3.
\section*{Appendix B. Additional results and analyses}
\textbf{Kernel Generation Variants.}
The dynamic convolution module (DCM), in which we generate dynamic kernels from the support foreground and employ convolution over the query feature, is an essential component in our proposed method. Therefore, here we presents two kernel generation variants: (i) we generate both asymmetric and symmetric kernel weights in parallel (ii) we first generate asymmetric kernel weights, and the asymmetric kernel weights are further used to generate symmetric kernel weights. With a kernel size 5, we term these two variants as $5/25$ (in parallel) and $5\rightarrow25$ (serial). As seen in Table \ref{A3}, these two variants achieve similar performance (66.0 \textit{vs} 66.7), which demonstrates the robustness of the DCM.
\setcounter{table}{0}  
\begin{table}[!t]
\centering
\renewcommand\arraystretch{1}
\resizebox{\columnwidth}{!}{
\begin{tabular}{c|cccccc}
\toprule[2pt]
\multicolumn{1}{c}{\multirow{2}{*}{Method}}      &  \multicolumn{5}{c}{1-shot mIoU}\\
 \multicolumn{1}{l}{} & \multicolumn{1}{l}{Fold0} & \multicolumn{1}{l}{Fold1} & \multicolumn{1}{l}{Fold2} & \multicolumn{1}{l}{Fold3} &
\multicolumn{1}{l}{Mean}& 
\multirow{-2}{*}{FB-IoU}\\ \hline
\multicolumn{1}{c}{5/25} & 65.9  & 70.6  & 66.9  & 60.5 & 66.0 &77.0\\
\multicolumn{1}{c}{5$\to$25}  & 65.7 & 71.6 & 69.1 & 60.6 & 66.7 &78.0\\
\bottomrule[2pt]
\end{tabular}}
\caption{Ablation studies for the kernel generation variants.}
\label{A3}
\end{table}
\begin{table}[!t]
\centering
\vspace{-1mm}
\renewcommand\arraystretch{1}
\resizebox{\columnwidth}{!}{
\begin{tabular}{c|ccccc}
\toprule[2pt]
\multicolumn{1}{c}{\multirow{2}{*}{Method}}      &  \multicolumn{5}{c}{1-shot mIoU}\\
 \multicolumn{1}{l}{} & \multicolumn{1}{l}{Fold0} & \multicolumn{1}{l}{Fold1} & \multicolumn{1}{l}{Fold2} & \multicolumn{1}{l}{Fold3} &
\multicolumn{1}{l}{Mean}\\ \hline
\multicolumn{1}{l}{PANet (Box) } & -  & -  & -  & - & 45.1\\
\multicolumn{1}{l}{CANet (Box) }  & - & - & - & - & 52.0\\
\multicolumn{1}{l}{Ours (Box) } & 59.8 & 70.5 & 63.2 & 55.5 & 62.3\\
\multicolumn{1}{l}{Ours (Pixel) } & 65.7  & 71.6  & 69.1  & 60.6 & 66.7\\
\bottomrule[2pt]
\end{tabular}}
\caption{Comparison with the existing methods under the bounding box supervision under 1-shot setting.}
\label{A2}
\end{table}
\begin{table}[!t]
\centering
\renewcommand\arraystretch{1}
\resizebox{\columnwidth}{!}{
\begin{tabular}{c|ccccc}
\toprule[2pt]

\multicolumn{1}{c}{\multirow{2}{*}{Method}}      & \multicolumn{1}{c}{\multirow{2}{*}{Backbone}} & \multicolumn{2}{c}{PASCAL-$5^i$}& \multicolumn{2}{c}{COCO-$20^i$}\\
\multicolumn{1}{c}{}  & \multicolumn{1}{c}{}  & \multicolumn{1}{l}{MIoU} & \multicolumn{1}{l}{FB-IoU} & \multicolumn{1}{l}{MIoU} & \multicolumn{1}{l}{FB-IoU}\\ \hline
\multicolumn{1}{c}{$w/o$ MS}                         & VGG16                                          & 61.3                 & 72.7                &  39.2            & 61.9                                   \\
\multicolumn{1}{l}{$w$ MS} & VGG16                                          & 61.7               & 73.7                    & 39.5               &  62.5                                    \\
\multicolumn{1}{c}{$w/o$ MS}                          & ResNet50                                       & 65.7                      &  77.4                   & 41.5                     &  62.7          \\
\multicolumn{1}{l}{$w$ MS} & ResNet50                                       &66.7                       & 78.0                     & 43.0                      & 63.2                                    \\ 

\bottomrule[2pt]
\end{tabular}}
\caption{Effectiveness of multi-scale testing under the 1-shot setting on the PASCAL-$5^i$ and the COCO-$20^i$ benchmarks.}
\label{A1}
\end{table}
\begin{table*}[!t]
\centering
\renewcommand\arraystretch{1}
\resizebox{\linewidth}{!}{
\begin{tabular}{c|cccccc|cccccc}
\toprule[2pt]

\multirow{2}{*}{Methods} &  \multicolumn{6}{c|}{1-shot}  & \multicolumn{6}{c}{5-shot}  \\
& \multicolumn{1}{l}{Fold-0} & Fold-1  & Fold-2  & Fold-3   & Mean  & FB-IoU
& \multicolumn{1}{l}{Fold-0} & Fold-1  & Fold-2  & Fold-3   & Mean  & FB-IoU \\ \hline
CANet (CVPR19) \cite{Zhang_2019_CVPR}                          
& 53.5  & 65.9  & 51.3  & 51.9   & 55.4  & 66.2                 
& 55.5  & 67.8  & 51.9  & 53.2   & 57.1  & 69.3           \\
CANet (CVPR19)+DCM 
& 64.7  & 66.8  & 51.8  & 51.9   & 58.8  & 69.3   
& 65.3  & 67.2  & 52.7  & 52.9   & 59.5  & 70.1            \\
PFENet (TPAMI’20) \cite{tian2020prior}
& 61.7  & 69.5  & 55.4  & 56.3   & 60.8  & 73.3 
& 63.1  & 70.7  & 55.8  & 57.9   & 61.9  & 73.9            \\
PFENet (TPAMI’20)+DCM
& 62.2  & 69.6  & 59.2  & 58.0   & 62.3  & 73.5 
& 63.1  & 70.0  & 60.0  & 58.5   & 62.9  & 73.6            \\

\bottomrule[2pt]
\end{tabular}}
\caption{Generalization ability of proposed DCM under both 1-shot and 5-shot settings.}
\label{A4}
\end{table*}
\begin{table*}[!t]
\centering
\renewcommand\arraystretch{1}
\resizebox{\linewidth}{!}{
\begin{tabular}{cc|cccccc|cccccc}
\toprule[2pt]

\multirow{2}{*}{Methods} & \multicolumn{1}{c|}{\multirow{2}{*}{Backbone}} & \multicolumn{6}{c|}{1-shot}  & \multicolumn{6}{c}{5-shot}  \\
& \multicolumn{1}{c|}{} 
& \multicolumn{1}{l}{Fold-0} & Fold-1  & Fold-2  & Fold-3   & Mean  & FB-IoU
& \multicolumn{1}{l}{Fold-0} & Fold-1  & Fold-2  & Fold-3   & Mean  & FB-IoU \\ \hline
Original Groudtruth \cite{Zhang_2019_CVPR}   & VGG16                        
& 58.9  & 69.1  & 63.2  & 55.7   & 61.7  & 73.7                 
& 63.4  & 70.7  & 68.1  & 59.0   & 65.3  & 77.2           \\
Non-original Groudtruth & VGG16 
& 59.3  & 69.5  & 63.3  & 55.8   & 62.0  & 73.8   
& 64.0  & 71.2  & 68.4  & 59.0   & 65.7  & 77.2            \\
Original Groudtruth \cite{tian2020prior}  &  ResNet50
& 65.7  & 71.6  & 69.1  & 60.6   & 66.7  & 78.0
& 70.0  & 73.2  & 70.9  & 65.5   & 69.9  & 80.7            \\
Non-original Groudtruth  &  ResNet50
& 65.6  & 71.8  & 69.2  & 60.5   & 66.8  & 78.0 
& 70.0  & 73.2  & 70.9  & 65.5   & 69.9  & 80.6            \\

\bottomrule[2pt]
\end{tabular}}
\caption{Comparison with Original Groudtruth and Non-original Groudtruth}
\label{A5}
\end{table*}
\textbf{Experiments with Bounding Box Annotations.}
Following PANet \cite{wang2019panet} and CANet \cite{Zhang_2019_CVPR}, we evaluate our model with weakly-supervised annotation (i.e, bounding box instead of pixel-wise annotation) on the support set. As shown in Table \ref{A2}, our method with the bounding box annotation achieves slightly inferior performance than that with expensive pixel-wise annotation. In addition, using bounding box annotation as supervision, our method also significantly outperforms both PANet and CANet. This experiment indicates the potential of our method in the segmentation task with weak supervision.

\textbf{Multi-Scale Testing.}
As a post-processing method, multi-scale testing\cite{zhao2021self} is widely adopted in many semantic segmentation tasks. Many few-shot semantic segmentation methods as well as our proposed method DPCN also use this strategy to improve segmentation performance. We present our results with / without multi-scale testing (termed as $w$ MS and $w/o$ MS, respectively) in Table \ref{A1}. The scales in our experiments are set as 473 and 641. We can see that (i) our model without the multi-scale testing also achieves state-of-the-art results using VGG16 and ResNet50 backbones, which further demonstrates the effectiveness of our proposed method (ii) multi-scale testing brings 1\% mIou improvement for our model with ResNet50 backbone, while a slight improvement when we use VGG16 as the backbone network.

\textbf{Generalization Ability of DCM.}
The dynamic convolution module (DCM) can  be used as a plug-and-play module to improve current prototype-based few-shot segmentation methods. We merge DCM into CANet and PFENet, and present the corresponding  results under both 1-shot and 5-shot settings in Table \ref{A4}. DCM further improve the performance of CANet and PFENet under both 1-shot and 5-shot settings, which shows the effectiveness of the DCM.

\textbf{Evaluation using Original Groundtruth.} As in PFENet \cite{tian2020prior}, we also evaluate our model with original groundtruth of the query image and the non-original one resized to the same size as training image size ($473\times473$). We can find from Table \ref{A5} that our proposed model obtains similar performance with the original or non-original query groundtruth.

\setcounter{figure}{0}  
\begin{figure}[!t]
	\centering
	\includegraphics[width=\columnwidth]{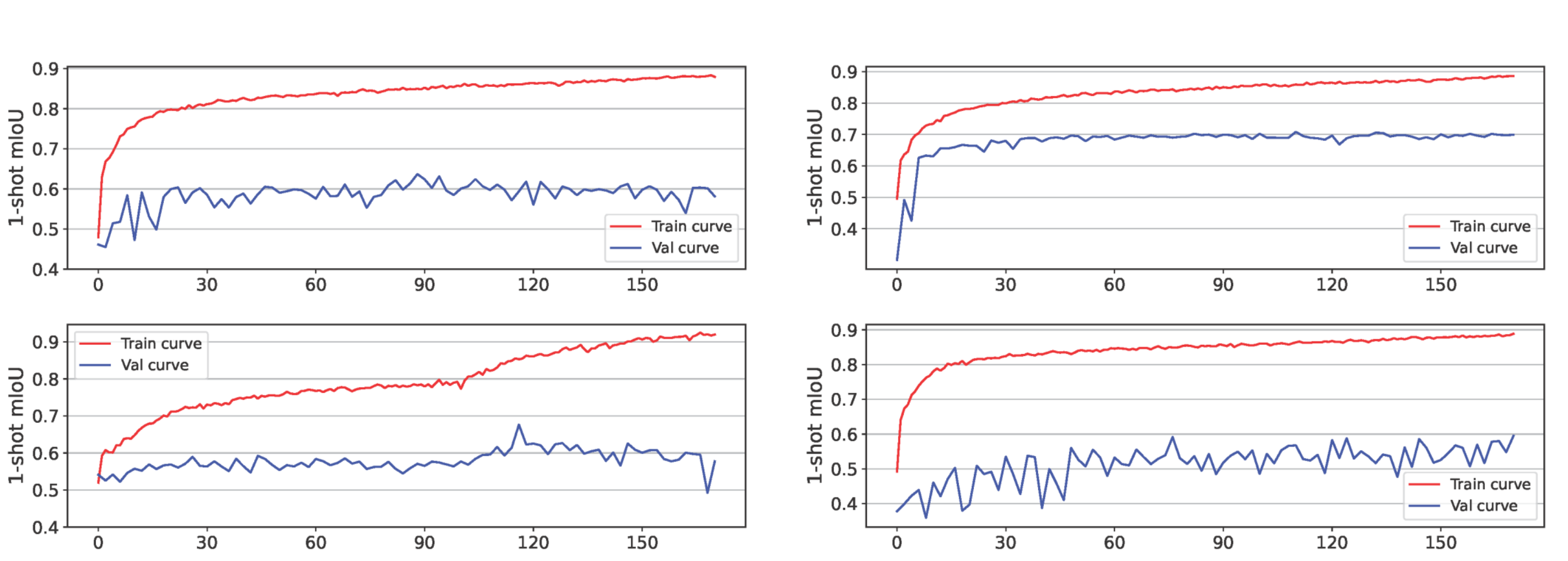}
	\caption{Training and Validation curves (x-axis: epochs, y-axis: 1-shot mIoU) on PASCAL-$5^i$ benchmark.}
	\label{F-A1}
\end{figure}
\begin{figure}[!t]
	\centering
	\includegraphics[width=\columnwidth]{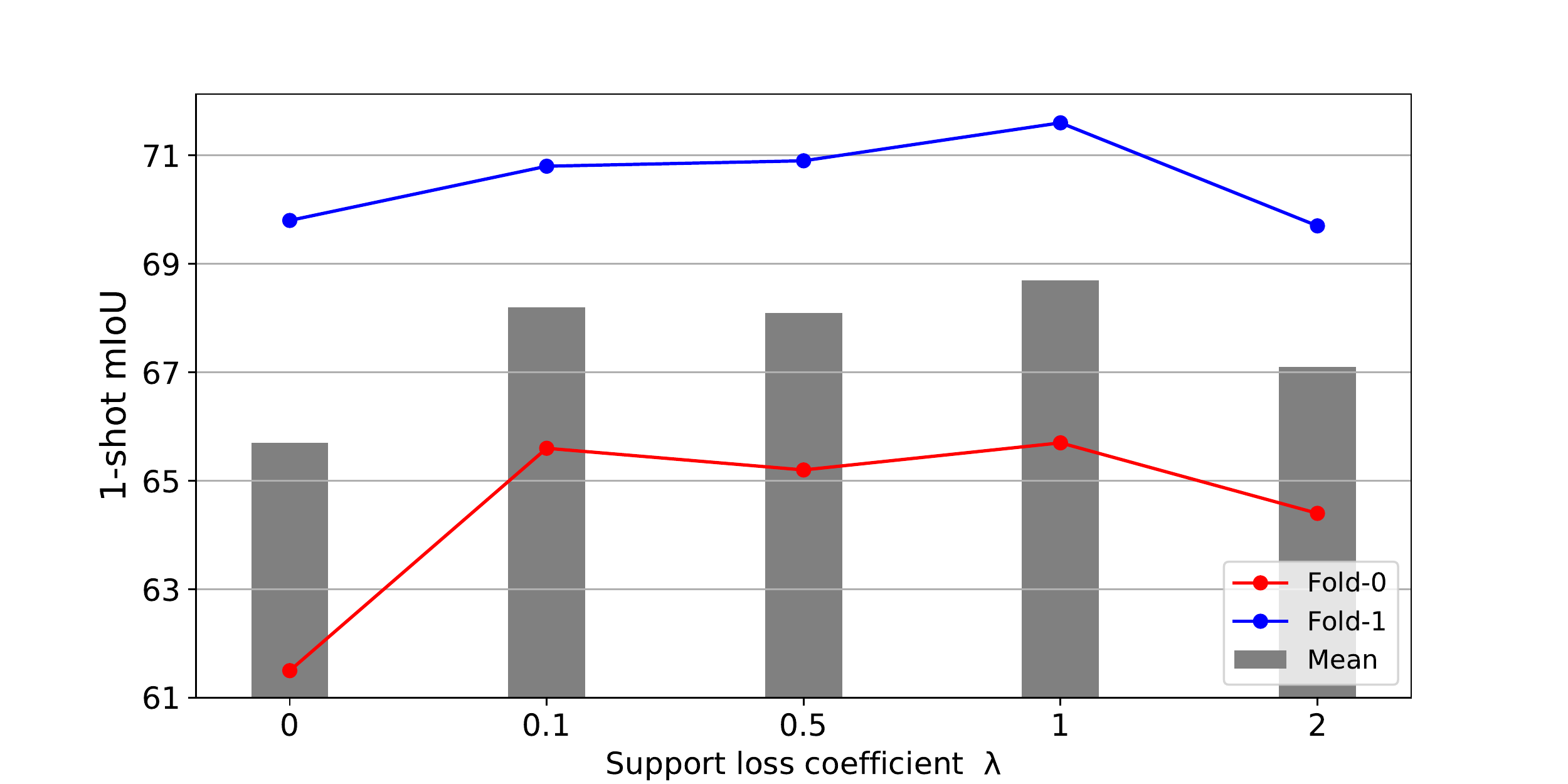}
	\caption{Ablation study for Support loss coefficient $\lambda$ on fold0 and fold1 of PASCAL-$5^i$ benchmark.}
	\label{F-A2}
\end{figure}

\textbf{Training and validation performance.} We present the performance (mIoU) changing process as the training epoch increases in Fig. \ref{F-A1}. As can be seen, the mIoU of the training process are much better than that of the validation in each fold. Besides, the validation mIoU in fold0 and fold1 are relatively stable, while the validation mIoU in fold2 and fold3 fluctuate as the training process goes on.

\textbf{Support Loss Coefficient $\lambda$.} During the model training, we use the predicted query mask as a pseudo mask for predicting the support mask, which requires a support loss $\mathcal{L}_{seg}^{q\rightarrow s}$ for supervision. For the support loss coefficient $\lambda$, we take its value from $\{0,0.1,0.5,1,2\}$ to study its influence on our model. The performance of the fold0 and fold1 as well as their mean on the PASCAL-$5^i$ benchmark are used for illustration. As shown in Fig. \ref{F-A2}, our model achieves best results when the support loss coefficient  is set as 1. And $\lambda$ is set as 1 in all our experiments.

\section*{Appendix C. Additional qualitative results}
\begin{figure*}[!t]
	\centering
	\includegraphics[width=\textwidth]{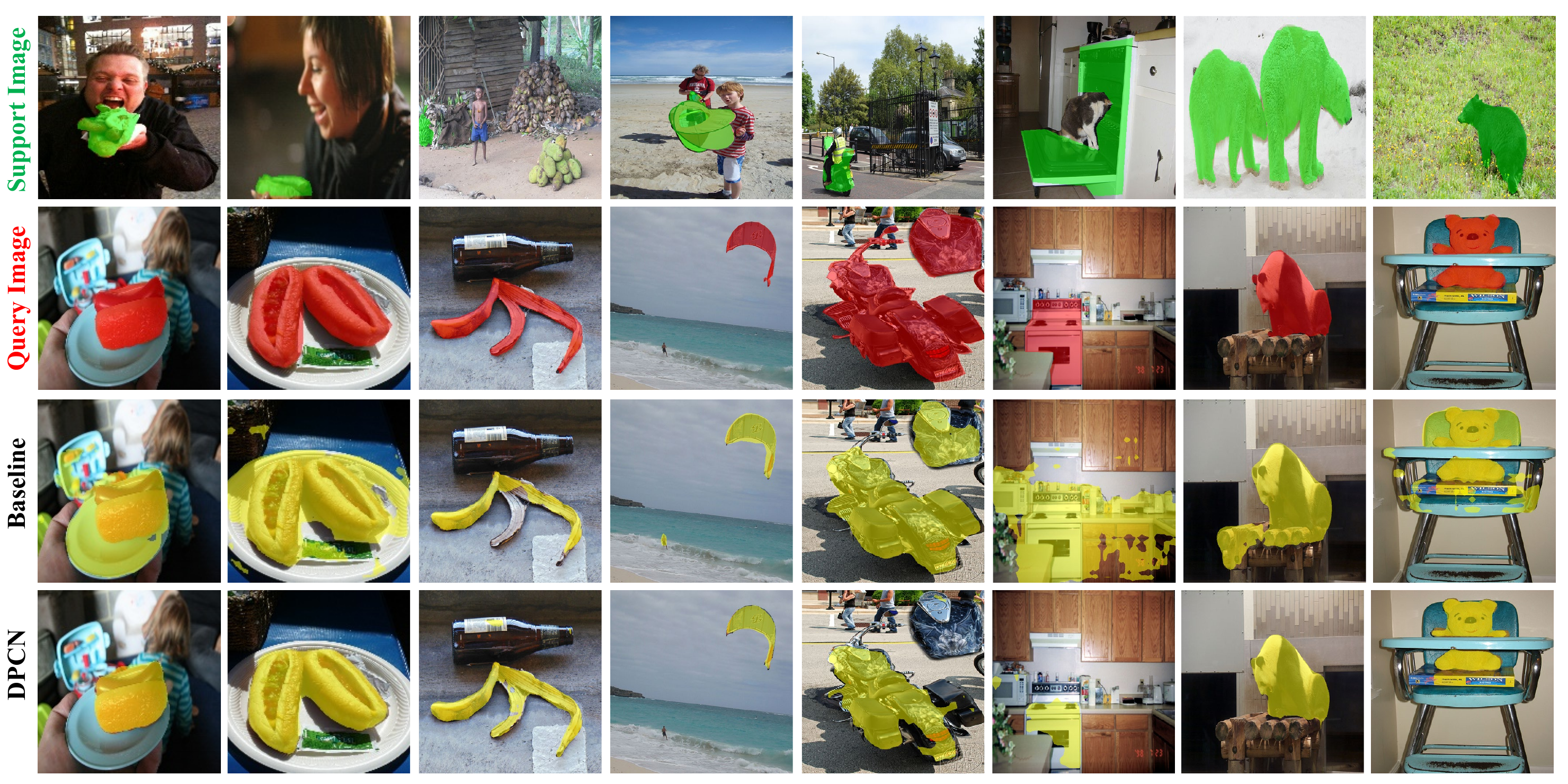}
	\caption{Qualitative results of our method DPCN and baseline model on COCO-$20^i$ benchmark with \textbf{large object appearance variations}. Zoom in for details.}
	\vspace{-6mm}
	\label{appearance}
\end{figure*}

\begin{figure*}[!t]
	\centering
	\includegraphics[width=\textwidth]{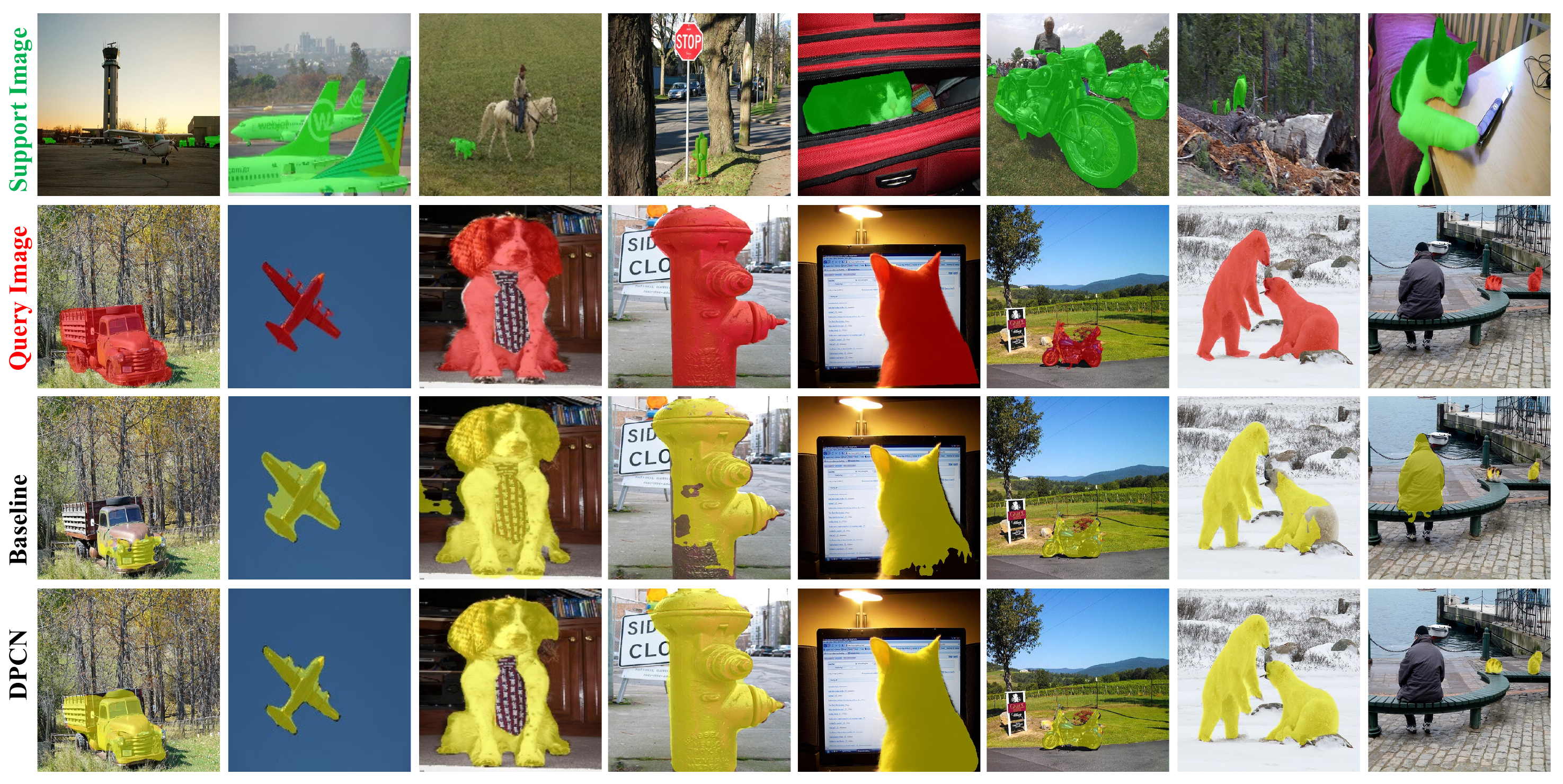}
	\caption{Qualitative results of our method DPCN and baseline model on COCO-$20^i$ benchmark with \textbf{large object scale variations}. Zoom in for details.}
	\label{scale}
\end{figure*}

In this section, we present more qualitative results of our proposed DPCN and its baseline to demonstrate its few-shot segmentation performance. Appearance and scale variations (more obvious in the COCO-$20^i$ benchmark) are the innate difficulty of the few-shot semantic segmentation task. Therefore, we first show some examples sampled from COCO-$20^i$ benchmark with large object appearance and scale variations in the Fig. \ref{appearance} and Fig. \ref{scale}, respectively. As can be seen, our model DPCN exhibits great superiority in alleviating appearance and scale variations. Besides, we also sample some examples from PASCAL-$5^i$ benchmark, and the qualitative results are presented in Fig. \ref{pascal}. Furthermore, DPCN occasionally predicts more accurate segmentation than human-annotated ground-truth (Fig. \ref{GT}), which further demonstrates the effectiveness of our method. Finally, we give some visualization of the support activation maps, initial pseudo mask as well as the refined pseudo mask in Fig. \ref{att}. We can see that the support activation maps can capture complementary object details in the query image, the initial pseudo mask gives rough pixel-wise location estimation of object, while the refined pseudo mask can estimate more accurate object location in the query image.

\begin{figure*}[!t]
	\centering
	\includegraphics[width=\textwidth]{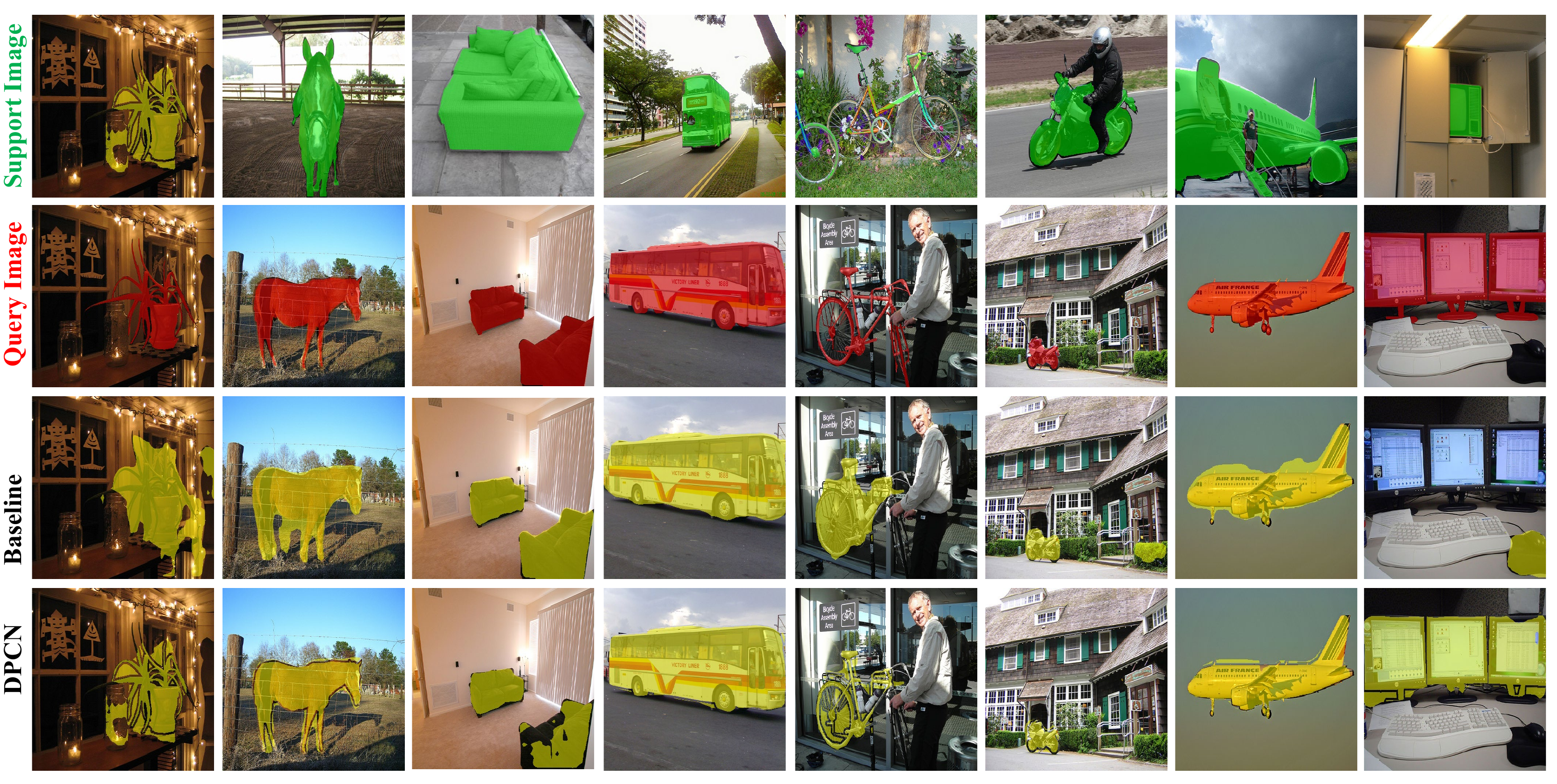}
	\caption{Qualitative results of our method DPCN and baseline model on PASCAL-$5^i$ benchmark. Zoom in for details.}
	\label{pascal}
	\vspace{-6mm}
\end{figure*}
\begin{figure*}[!t]
	\centering
	\includegraphics[width=\textwidth]{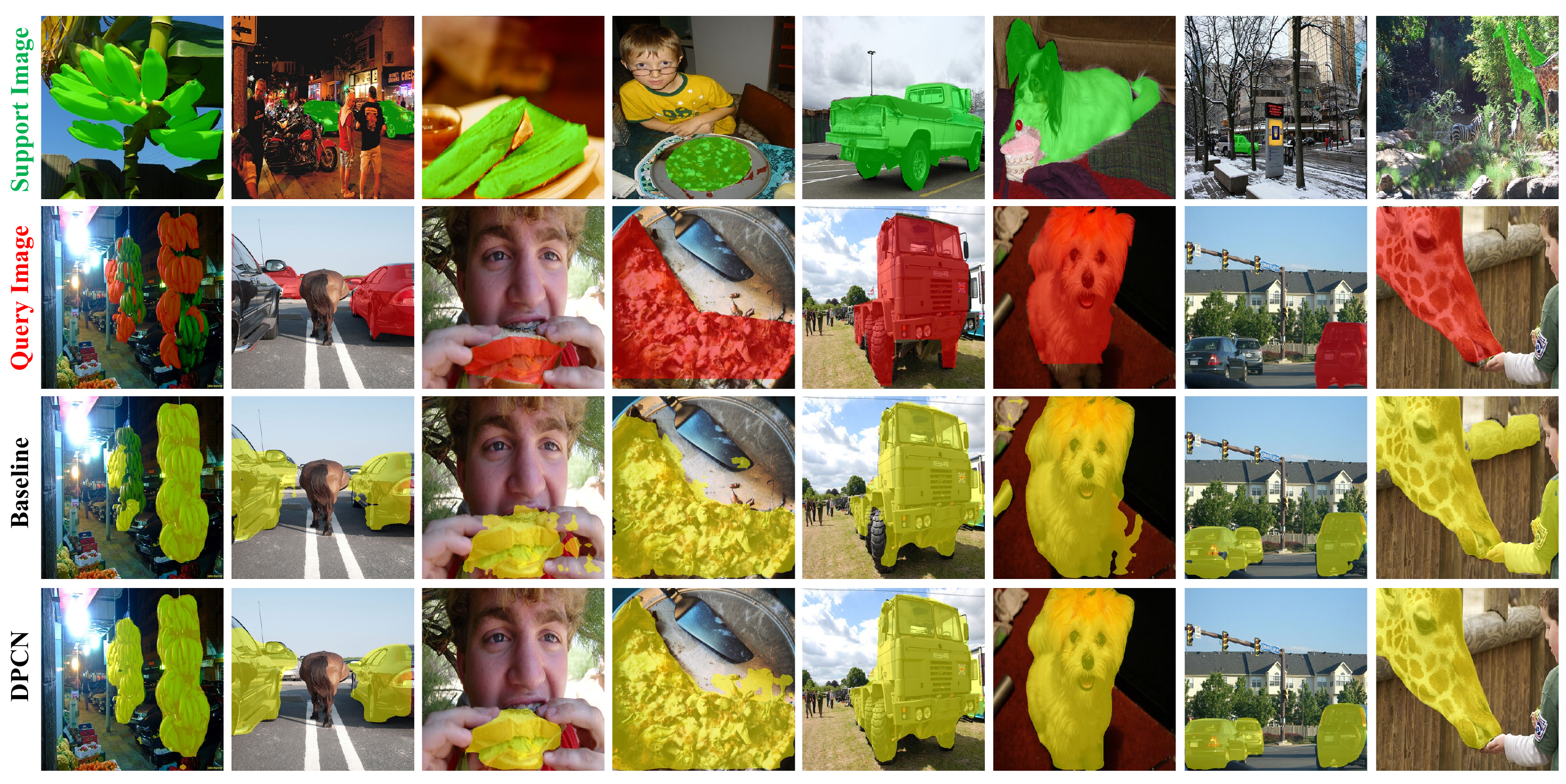}
	\caption{Our proposed DPCN occasionally predicts more accurate segmentation
masks than human-annotated ground-truths. Examples are sampled from both PASCAL-$5^i$ and COCO-$20^i$. Zoom in for details.}
	\label{GT}
\end{figure*}
\begin{figure*}[!t]
	\centering
	\includegraphics[angle=270,scale=0.7]{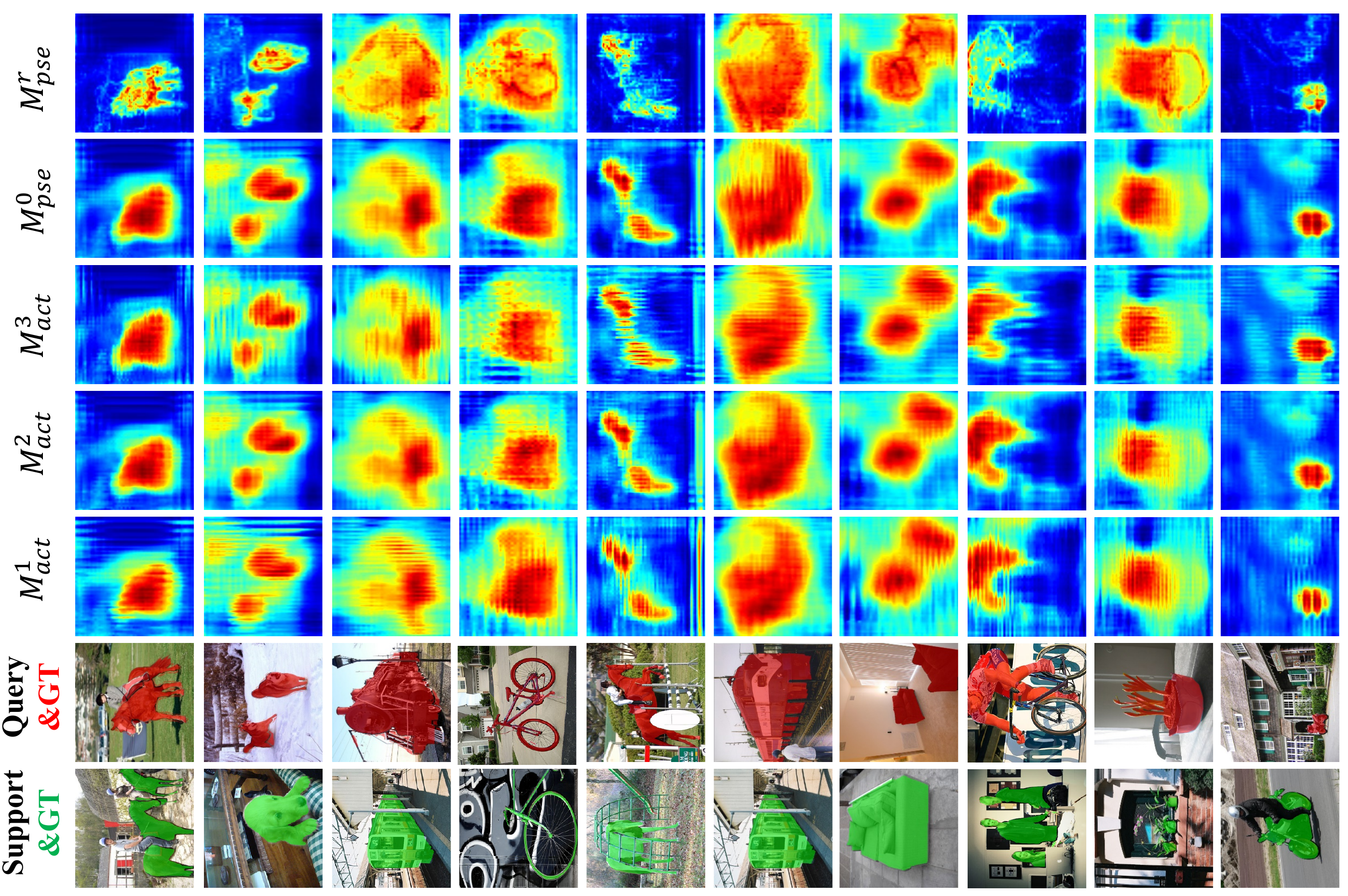}
	\caption{Visualization of the support activation maps $\{M_{act}^{i}\}_{i=1}^3$ and the initial pseudo mask $M_{pse}^0$ in the support activation module (SAM), as well as the refined pseudo mask $M_{pse}^r$in the feature filtering module (FFM). Zoom in for details.}
	\label{att}
\end{figure*}

\end{document}